\makeatletter\def\graphicscache@inhibit{true}\makeatother

\documentclass[letterpaper, 10 pt, conference]{ieeeconf}  %

\IEEEoverridecommandlockouts                              %

\usepackage{amsmath,amssymb,amsfonts}
\usepackage{pifont}
\usepackage{textcomp}
\usepackage{multirow}
\usepackage[hang,flushmargin,symbol]{footmisc}
\usepackage{pgfplots}
\pgfplotsset{compat=newest}
\usepackage{tikz}

\usepackage{algorithm}
\usepackage{algpseudocode}

\usetikzlibrary{fit,arrows,arrows.meta,automata,backgrounds,calc,chains,%
decorations.markings,decorations.pathreplacing,decorations.pathmorphing,%
matrix,positioning,shapes,shapes.geometric,shapes.symbols,spy,trees,tikzmark}

\pgfdeclarelayer{foreground}
\pgfsetlayers{background,main,foreground}

\newcommand{\relcoord}[4]{%
    \path let \p1=(#1.south west), \p2=(#1.north east) in
    coordinate (#2) at ({\x1+#3*(\x2-\x1)},{\y1+#4*(\y2-\y1)});
}

\newcommand{\zoomconnect}[4]{%
    \path let \p1=(#1), \p2=(#2) in \pgfextra{
        \pgfmathsetmacro{\zdx}{(\x2-\x1)/1pt}
        \pgfmathsetmacro{\zdy}{(\y2-\y1)/1pt}
        \pgfmathsetmacro{\zd}{veclen(\zdx,\zdy)}
        \pgfmathsetlengthmacro{\zrAdim}{#3}
        \pgfmathsetlengthmacro{\zrBdim}{#4}
        \pgfmathsetmacro{\zrA}{\zrAdim/1pt}
        \pgfmathsetmacro{\zrB}{\zrBdim/1pt}
        \pgfmathsetmacro{\zphi}{atan2(\zdy,\zdx)}
        \pgfmathsetmacro{\zbeta}{acos((\zrA-\zrB)/\zd)}
        \xdef\ztangentA{\zphi+\zbeta}
        \xdef\ztangentB{\zphi-\zbeta}
    };
    \draw[white,line width=1.2pt,shorten <=2.4pt,shorten >=2.4pt] ($(#1)+(\ztangentA:#3)$) -- ($(#2)+(\ztangentA:#4)$);
    \draw[white,line width=1.2pt,shorten <=2.4pt,shorten >=2.4pt] ($(#1)+(\ztangentB:#3)$) -- ($(#2)+(\ztangentB:#4)$);
}

\newcommand{\zoomspy}[4]{%
    \pgfmathsetlengthmacro{\zoomradius}{0.5*#4}
    \pgfmathsetlengthmacro{\sourceradius}{0.5*#4/#3}
    \zoomconnect{#1}{#2}{\sourceradius}{\zoomradius}
    \spy[magnification=#3,size=#4] on (#1) in node at (#2);
}

\usepackage{hyperref}
\usepackage{url}
\usepackage{graphicx}
\usepackage{booktabs}
\usepackage{acronym}
\usepackage[capitalize]{cleveref}
\usepackage{siunitx}
\usepackage{cite}
\usepackage{flushend}
\usepackage{microtype}

\usepackage{multirow}
\usepackage{gensymb}

\DeclareMathOperator*{\argmin}{arg\,min}

\algrenewcommand\algorithmicrequire{\textbf{Input:}}
\algrenewcommand\algorithmicensure{\textbf{Output:}}

\acrodef{FoV}{Field of View}
\acrodef{LiDAR}{Light Detection and Ranging}
\title{\LARGE \bf
Generalizable Target-Less Joint Intrinsic--Extrinsic Camera--LiDAR Calibration using Deep Pixel--Point Correspondences
}

\author{Simon Bultmann, Daniele Cattaneo, and Abhinav Valada%
\thanks{Department of Computer Science, University of Freiburg, Germany}%
\thanks{This work was funded by the German Research Foundation (DFG) Emmy Noether Program under grant 468878300.}%
}

\begin{document}

\maketitle
\thispagestyle{empty}
\pagestyle{empty}

\begin{abstract}
Accurate camera-LiDAR calibration is an essential prerequisite for robust multi-modal perception in robotics.
Recent target-less methods based on learned pixel-point correspondences achieve high extrinsic calibration accuracy and generalize across environments.
However, they operate on rectified images with known camera intrinsics, restricting their applicability to systems whose cameras have already been individually calibrated.
We present a fully automatic, target-less framework that jointly estimates camera intrinsics, radial-tangential distortion, and camera-LiDAR extrinsics from raw images and LiDAR scans.
Our method generalizes deep correspondence-based calibration from rectified images to raw cameras through intrinsic- and distortion-aware correspondence learning via synthetic augmentations.
During calibration, we iteratively couple correspondence prediction and nonlinear optimization to jointly refine the camera parameters and relative sensor transformation.
We evaluate our approach on an unseen camera-LiDAR configuration from the KITTI dataset and on three real robotic platforms in a zero-shot manner.
The results demonstrate that joint calibration with our generalized pixel-point correspondence network improves extrinsic accuracy over correspondence-based methods using pre-calibrated cameras, while additionally recovering accurate intrinsics.
\end{abstract}

\section{Introduction}
Accurate camera-LiDAR calibration is a fundamental requirement for robotic systems that fuse complementary visual and geometric information for robust perception in real-world environments~\cite{valada2016towards,bultmann2021real,mohan2026up,hindel2026cotrate}.
To this end, measurements from both modalities must be related, e.g., by projecting LiDAR points into the camera image, requiring accurate camera parameters and a rigid six degree-of-freedom (6~DoF) relative sensor transformation. Calibration errors lead to spatially inconsistent measurements and can directly degrade downstream perception performance.

Camera-LiDAR calibration has traditionally been performed using specially designed targets and controlled acquisition procedures.
These include checkerboards or fiducial markers for camera intrinsic calibration~\cite{rehder_kalibr_2016}, hybrid visual-geometric targets that can be detected by both cameras and LiDARs~\cite{kuemmerle_unified_2020,yan_joint_2023}, and larger structured calibration environments~\cite{wiesmann_joint_2024}.
Although target-based approaches achieve high accuracy, they are labor-intensive and require dedicated equipment and expert supervision. The need to repeat the procedure after sensor replacement, vibration, mechanical stress, or thermal effects makes them difficult to scale across large robotic systems or fleets.

To overcome these limitations, recent research has focused on target-less calibration using naturally occurring scene structure.
Existing approaches derive calibration constraints from semantic objects or keypoints~\cite{paetzold_camcalib_2022}, geometric primitives such as edges and planes~\cite{li_joint_2023}, sensor motion~\cite{petek_automatic_2024}, or learned multi-modal correspondences~\cite{cattaneo2020cmrnetpp,cattaneo_cmrnext_2025,nayak2024ralf}.
For camera-LiDAR matching, optical-flow-based methods such as I2DLoc~\cite{yu_i2dloc_2024} and CMRNext~\cite{cattaneo_cmrnext_2025} currently set the state of the art.
These methods formulate pixel-point matching as dense displacement estimation between projected LiDAR scans and RGB images. Building on this paradigm, MDPCalib~\cite{petek_automatic_2024} combines motion-based initialization with learned correspondences to obtain a fully automatic target-less extrinsic calibration pipeline. However, existing deep correspondence-based methods operate on rectified images with known camera intrinsics and therefore require calibrating each camera beforehand.

\begin{figure}[t]
\centering

\resizebox{\linewidth}{!}{%
\begin{tikzpicture}[x=1cm,y=1cm,>=Latex]

\tikzset{
  panel/.style={
    rounded corners=0.35cm,
    draw=black!25,
    line width=0.6pt,
    fill=black!1
  },
  title/.style={
    font=\bfseries\fontsize{18}{20}\selectfont
  },
  subtitle/.style={
    font=\bfseries\fontsize{15}{17}\selectfont
  },
  sensorcircle/.style={
    draw=red!85!black,
    line width=2.pt
  },
  axisx/.style={-Latex, red, line width=0.9pt},
  axisy/.style={-Latex, green!60!black, line width=0.9pt},
  axisz/.style={-Latex, blue, line width=0.9pt},
  midarrow/.style={-Latex, line width=1.0pt, draw=black!70},
}

\newcommand{\cameraaxes}[1]{%
  \begin{scope}
    \coordinate (O) at #1;
    \draw[axisx] (O) -- ++(-0.55,-0.65) node[below] {$x$};
    \draw[axisy] (O) -- ++(0.00,0.95) node[above] {$y$};
    \draw[axisz] (O) -- ++(0.95,-0.10) node[right] {$z$};
  \end{scope}
}

\newcommand{\lidaraxes}[1]{%
  \begin{scope}
    \coordinate (O) at #1;
    \draw[axisx] (O) -- ++(1.05,0.00) node[right] {$x$};
    \draw[axisy] (O) -- ++(0.00,0.95) node[above] {$y$};
    \draw[axisz] (O) -- ++(-0.38,-0.82) node[below] {$z$};
  \end{scope}
}

\draw[panel] (0.2,7.25) rectangle (15.8,13.25);
\draw[panel] (0.2,2.40) rectangle (15.8,7.05);

\node[subtitle]    at (8.0,12.85) {Robots with multimodal sensors};
\node[subtitle] at (8.0,6.55)  {Automatic target-less joint intrinsic and extrinsic calibration};

\node[inner sep=0pt] (car)   at (3.0,10.25)
  {\includegraphics[height=5.75cm]{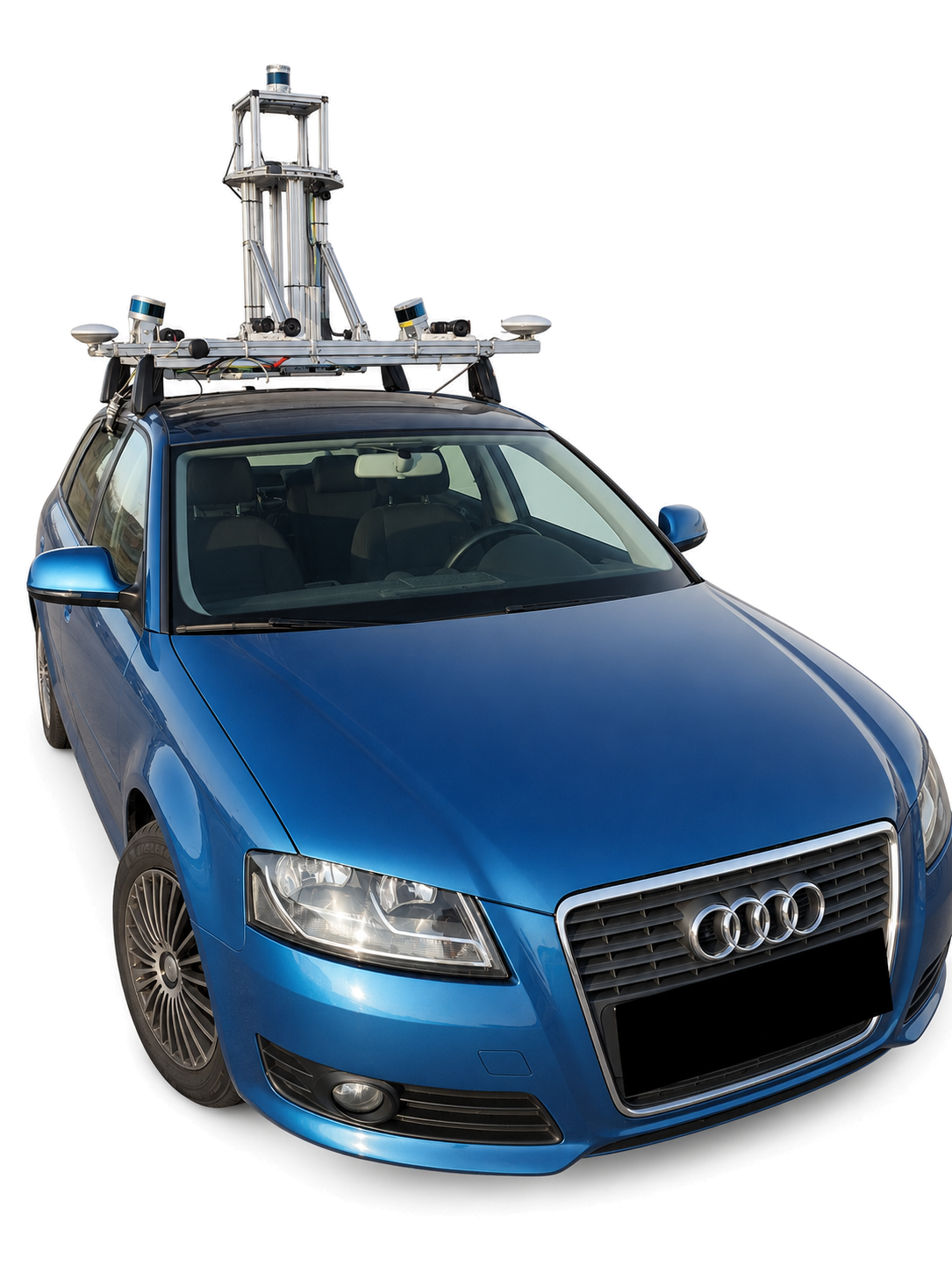}};

\node[inner sep=0pt] (spot)  at (7.5,9.85)
  {\includegraphics[height=5.25cm]{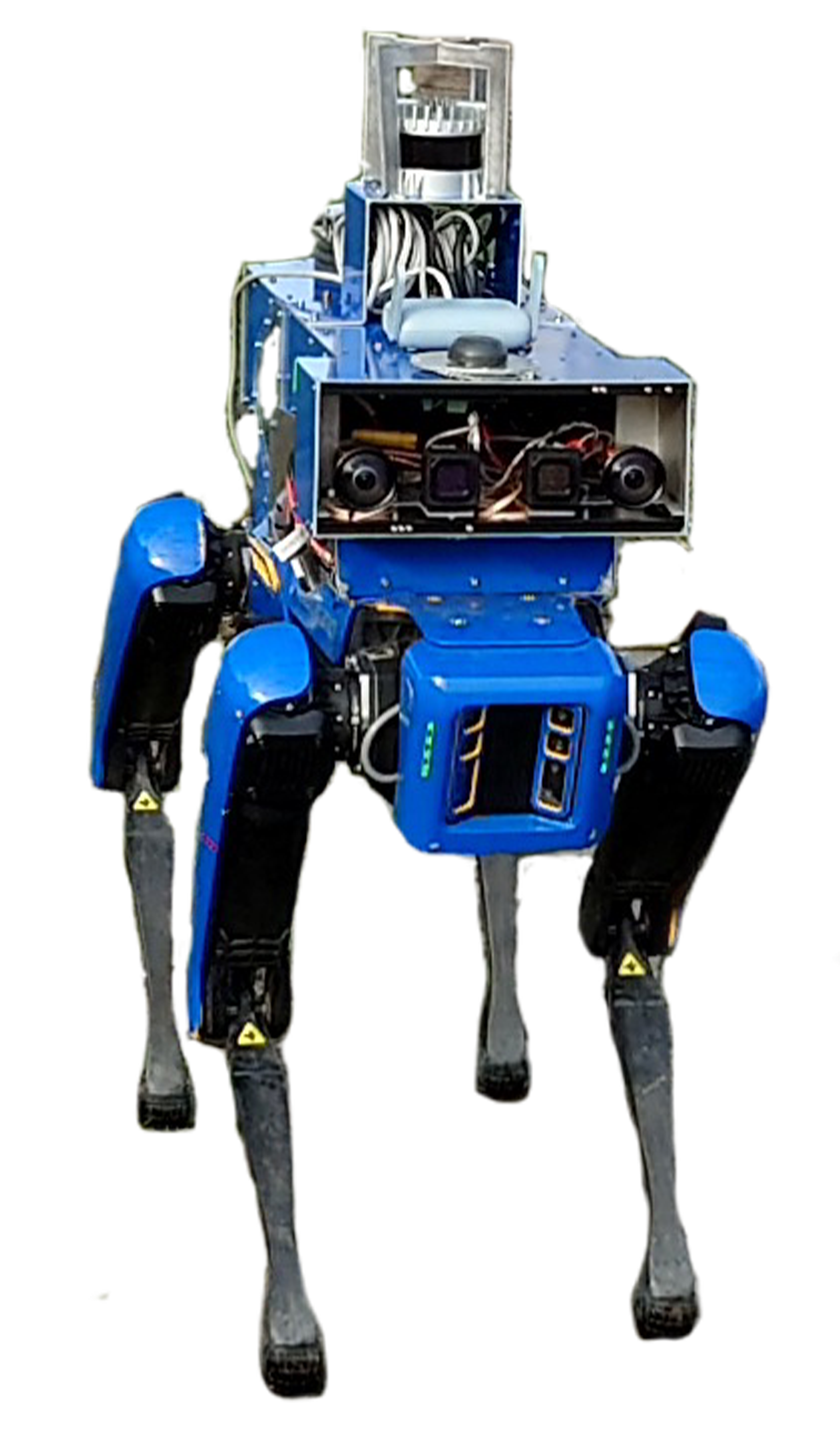}};

\node[inner sep=0pt] (drone) at (12.7,9.85)
  {\includegraphics[height=5.65cm]{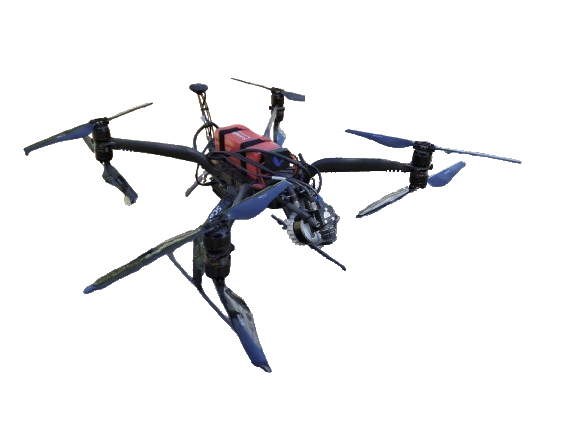}};

\draw[sensorcircle] ($(car.north)+(-.90,-0.35)$) circle[radius=0.28];
\draw[sensorcircle] ($(car.north west)+(1.25,-1.45)$) circle[radius=0.28];
\draw[sensorcircle] ($(car.north west)+(2.05,-1.45)$) circle[radius=0.28];

\draw[sensorcircle] ($(spot.north)+(0.10,-0.60)$) circle[radius=0.33];
\draw[sensorcircle] ($(spot.center)+(0.25, 0.90)$) ellipse[x radius=0.9cm,y radius=0.38cm];

\draw[sensorcircle] ($(drone.center)+(0.35,-0.15)$) circle[radius=0.4];

\node[font=\Large] at (0.8,5.30) {$\mathbf{K}$};
\node[inner sep=0pt] (intr) at (1.85,5.30)
  {\includegraphics[height=1.25cm]{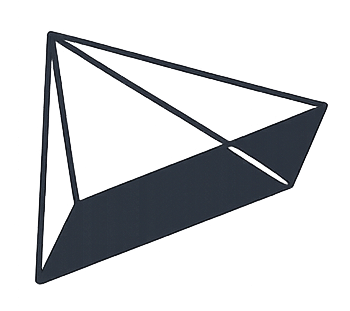}};

\node[font=\Large] at (0.8,3.50) {$\mathbf{D}$};
\node[inner sep=0pt] (dist) at (1.85,3.50)
  {\includegraphics[height=1.25cm]{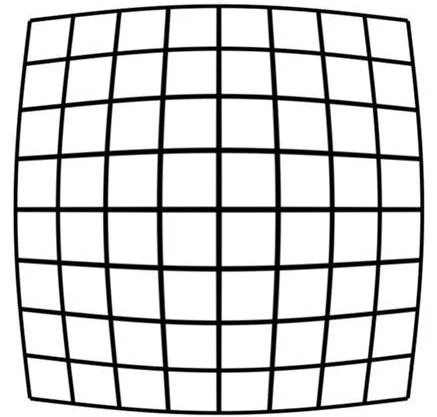}};

\draw[decorate,line width=2pt,decoration={brace,amplitude=10pt}]
  (2.65,6.0) -- (2.65,2.65);

\node[inner sep=0pt] (cam) at (5.55,4.0)
  {\includegraphics[height=2.45cm]{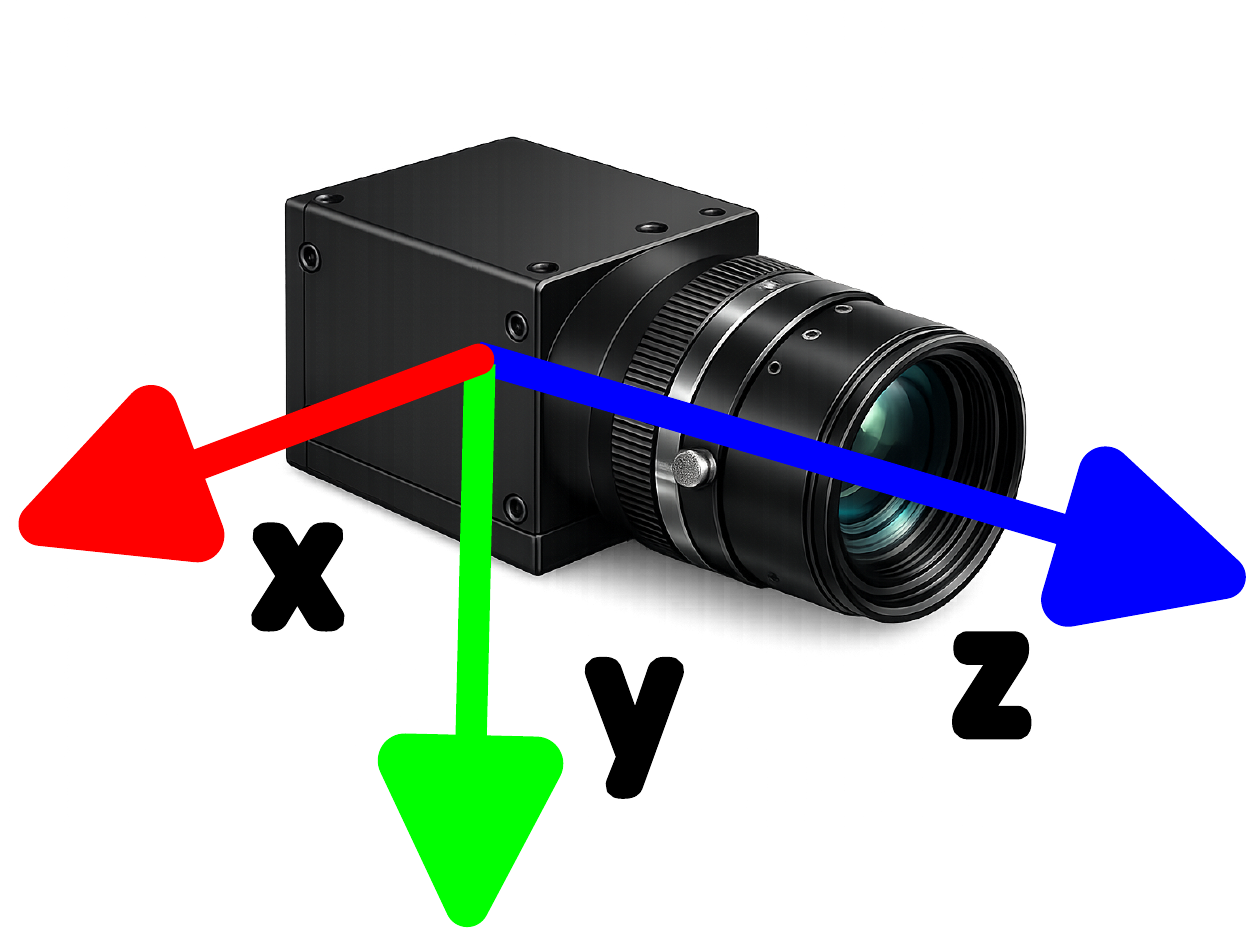}};

\node[inner sep=0pt] (lidar) at (12.05,4.0)
  {\includegraphics[height=4.35cm]{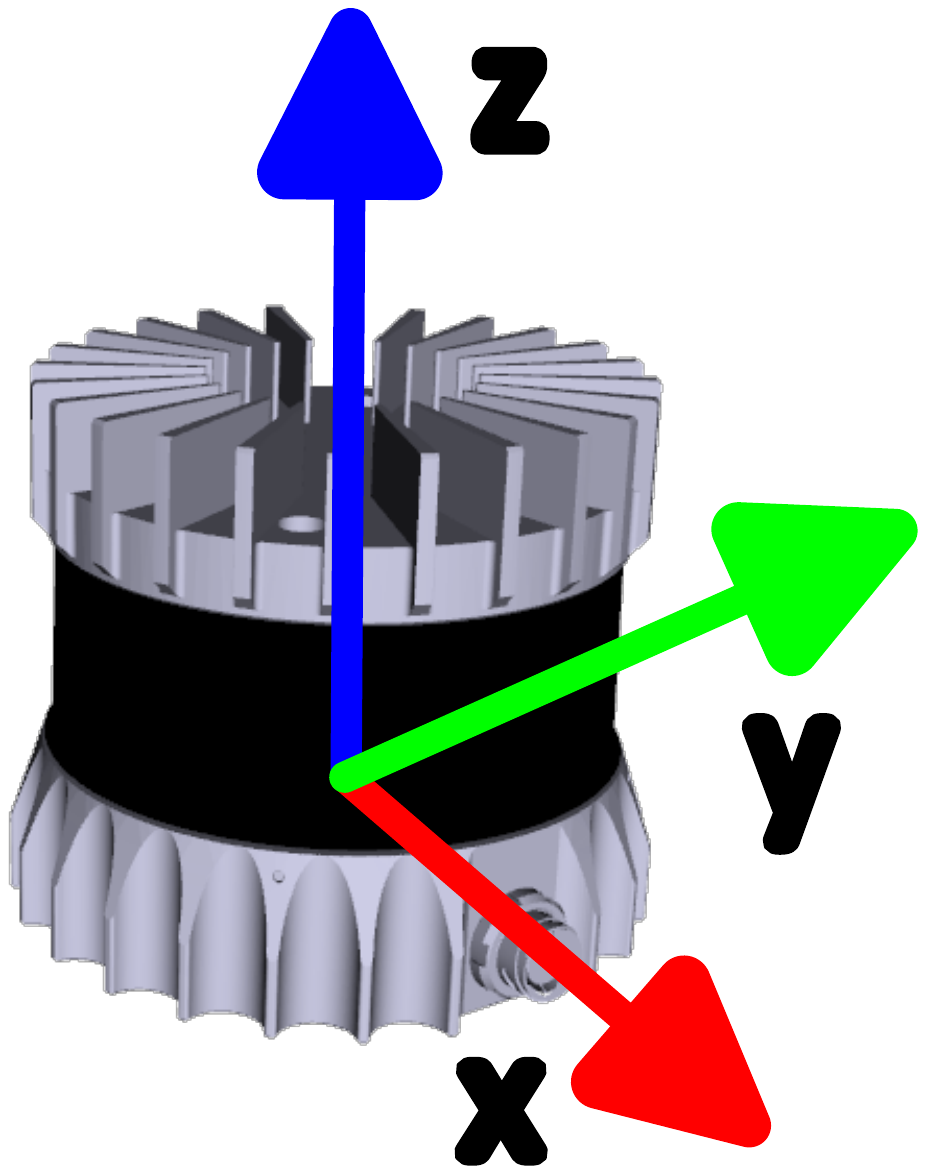}};

\draw[dashed, line width=2.pt, <->]
  (6.8,4.10) .. controls (8.10,4.95) and (9.85,4.95) .. (11.2,4.10);

\node[font=\Large] at (9.,5.2) {$\mathbf{T}_{\mathrm{calib}}$};

\end{tikzpicture}
}
\vspace{-2.7em}
\caption{\textbf{Fully automatic, target-less joint intrinsic and extrinsic camera-LiDAR calibration:} Our method generalizes to diverse robotic platforms and jointly estimates camera intrinsics $\mathbf{K}$, radial-tangential distortion $\mathbf{D}$, and the camera-LiDAR extrinsic transformation $\mathbf{T}_{\mathrm{calib}}$.}
\vspace{-1.9em}
\label{fig:teaser}
\end{figure}

We bridge this gap with a fully automatic, target-less framework that jointly estimates camera intrinsics, radial-tangential distortion, and camera-LiDAR extrinsics from raw images and LiDAR scans across diverse robotic platforms, as illustrated in Fig.~\ref{fig:teaser}.
Our method generalizes deep correspondence-based calibration from rectified images to raw, distorted cameras through intrinsic- and distortion-aware correspondence learning with physically plausible camera-parameter augmentation.
During calibration, we use the current intrinsic and extrinsic estimates to project the LiDAR scan into the image and provide the network input for pixel-point correspondence prediction. The predicted correspondences then jointly update the calibration parameters via nonlinear optimization, providing the estimates for the next projection. This iterative feedback loop progressively recovers both the camera parameters and the relative sensor transformation.

Generalization across sensor configurations is particularly important for practical learning-based calibration. Common evaluation on an \emph{unseen sequence} from the same rigidly mounted camera-LiDAR pair does not establish generalization to an \emph{unseen sensor configuration}, since the camera intrinsics and sensor transformation remain unchanged and can lead to overfitting to sensor-specific projection patterns~\cite{huang_what_2025}. We mitigate this problem by training across multiple datasets and sensor setups, separating learned correspondence estimation from geometric optimization, and augmenting camera parameters, while focusing our evaluation on camera-LiDAR configurations not represented during training.

In summary, our main contributions are:
\begin{enumerate}
\item A fully automatic, target-less framework for jointly estimating camera intrinsics, radial-tangential distortion, and camera-LiDAR extrinsics from raw sensor observations.
\item Intrinsic- and distortion-aware pixel-point correspondence learning with physically plausible camera-parameter augmentation, enabling a single network to operate across diverse raw camera configurations.
\item An iterative calibration procedure that couples correspondence prediction with joint nonlinear optimization.
\item Extensive evaluations comprising an unseen camera-LiDAR configuration from the KITTI dataset and zero-shot transfer to three real robotic platforms without sensor-specific retraining.
\item We will release the code and pre-trained models upon acceptance. 
\end{enumerate}

\section{Related Work}

Camera-LiDAR calibration methods fall into target-based and target-less approaches.
Target-based methods use calibration objects observable in both modalities, e.g., planar or hybrid visual-geometric targets or larger structured environments~\cite{kuemmerle_unified_2020,yan_joint_2023,wiesmann_joint_2024}.
They provide accurate constraints and often support joint intrinsic-extrinsic estimation, but require dedicated equipment and calibration-specific data collection.

Target-less methods instead exploit information naturally available during robot operation.
Classical approaches use geometric scene structure, hand-crafted similarity measures, or sensor motion.
Yuan~et~al.~\cite{yuan_pixel_2021} and Yin~et~al.~\cite{yin_automatic_2023} refine extrinsics through edge alignment, while Koide~et~al.~\cite{koide_general_2023} combine automatic initialization with normalized-information-distance registration.
Motion-based approaches align visual and LiDAR trajectories~\cite{yin_automatic_2023}, but require informative motion and subsequent refinement.
These methods avoid fiducial targets, but estimate only extrinsics and rely on scene-specific geometric or appearance assumptions.

Learning-based methods replace manually designed similarities with learned cross-modal representations.
RegNet~\cite{schneider_regnet_2017}, CalibNet~\cite{iyer_calibnet_2018}, and LCCNet~\cite{lv_lccnet_2021} directly regress the 6~DoF transformation from RGB images and projected LiDAR data.
While accurate when trained and evaluated on the same rigid sensor setup, direct regression depends strongly on fixed camera intrinsics and sensor-specific projection patterns, limiting generalizability to unseen sensor configurations~\cite{huang_what_2025}.

Correspondence-driven approaches use learning for cross-modal matching while recovering sensor poses through explicit geometry.
I2DLoc~\cite{yu_i2dloc_2024} and CMRNext~\cite{cattaneo_cmrnext_2025} formulate pixel-point association as dense displacement estimation between RGB images and projected LiDAR measurements, using the resulting matches for PnP-based camera pose estimation in pre-built LiDAR maps. 
CMRNext further extends this formulation to extrinsic camera-LiDAR calibration and generalizes to sensor configurations unseen during training. %
MDPCalib~\cite{petek_automatic_2024} combines learned correspondences with motion-based initialization and nonlinear optimization to obtain a fully automatic target-less extrinsic calibration pipeline.
However, these methods operate on rectified images 
and therefore address only extrinsic calibration.

Recent scene-based methods estimate camera-LiDAR extrinsics through differentiable neural rendering rather than explicit correspondences.
MOISST~\cite{herau_moisst_2023} and SOAC~\cite{herau_soac_2024} jointly optimize neural radiance fields with sensor parameters, while subsequent methods use LiDAR-supported Gaussian representations~\cite{herau_3dgscalib_2024,zhou_robust_2025,jung_targetless_2025}.
These methods provide a complementary scene-level formulation, but assume known camera intrinsics and primarily address extrinsic calibration. We instead build on correspondence-based calibration, whose separation of learned cross-modal matching from explicit geometric optimization facilitates joint intrinsic and extrinsic calibration while retaining cross-sensor generalization.

Only a few target-less methods estimate both camera intrinsics and camera-LiDAR extrinsics.
Li~et~al.~\cite{li_joint_2023} use plane-constrained bundle adjustment that couples visual reconstruction with LiDAR plane registration to jointly estimate camera intrinsics, including radial distortion, and camera-LiDAR extrinsics. However, they evaluate joint performance only on an internal dataset, whereas the KITTI~\cite{kitti_geiger_2012} experiments use ground-truth camera intrinsics.
Shen~et~al.~\cite{shen_direct_2025} alternate intrinsic and extrinsic refinement using normalized information distance between camera and LiDAR intensities.
Distortion is estimated during coarse initialization, while the joint stage refines only the pinhole parameters and extrinsics. The evaluation is performed on several sensor suites but does not include a standard benchmark such as KITTI.

Existing target-less joint approaches therefore rely on geometric primitives or direct intensity alignment, often using limited camera models or evaluation settings.
In contrast, our framework combines generalizable learned cross-modal matching with nonlinear optimization.
We learn intrinsic- and distortion-aware pixel-point correspondences from raw images and alternate correspondence prediction with joint refinement of the full pinhole model with radial-tangential distortion and the 6~DoF rigid sensor transformation.

\section{Method}
\label{sec:method}
\begin{figure*}[t]
\centering
\resizebox{\textwidth}{!}{%
\begin{tikzpicture}[
    x=1cm,y=1cm,>=Latex,
    panel/.style={
        rounded corners=3mm,
        draw=blue!35!black,
        line width=.8pt,
        fill=blue!2
    },
    head/.style={
        rounded corners=1.5mm,
        fill=blue!45!black,
        text=white,
        font=\sffamily\bfseries\small,
        inner xsep=5pt,
        inner ysep=2.5pt
    },
    stage/.style={
        rounded corners=2mm,
        draw=orange!65!black,
        line width=.7pt,
        fill=orange!7
    },
    stitle/.style={
        font=\sffamily\bfseries\scriptsize,
        align=center
    },
    tinylabel/.style={
        font=\sffamily\scriptsize,
        align=center
    },
    flow/.style={
        -{Latex[length=1.8mm,width=1.3mm]},
        line width=.9pt,
        draw=black!85
    },
    datacam/.style={
        -{Latex[length=1.55mm,width=1.15mm]},
        line width=.8pt,
        draw=blue!72!black
    },
    datalidarX/.style={
        line width=.8pt,
        draw=cyan!55!blue!75!black
    },
    datalidar/.style={
        -{Latex[length=1.55mm,width=1.15mm]},
        line width=.8pt,
        draw=cyan!55!blue!75!black
    },
    feedback/.style={
        -{Latex[length=1.8mm,width=1.3mm]},
        line width=.9pt,
        dashed,
        draw=black
    },
    trajvis/.style={
        draw=teal!75!black,
        line width=1.1pt,
        postaction={decorate},
        decoration={
            markings,
            mark=between positions .18 and .90 step .18
            with {\arrow{Latex}}
        }
    },
    trajlidar/.style={
        draw=green!55!black,
        line width=1.1pt,
        postaction={decorate},
        decoration={
            markings,
            mark=between positions .18 and .90 step .18
            with {\arrow{Latex}}
        }
    }
]
\begin{scope}[on background layer]
    \draw[panel] (0,1.05) rectangle (1.70,5.15);
    \draw[panel] (2.05,1.05) rectangle (8.25,5.15);
    \draw[panel] (8.85,0.6) rectangle (20.75,5.15);
    \draw[panel] (21.00,1.4) rectangle (24.05,4.45);
\end{scope}

\node[head] at (.85,5.15) {(a) Input};
\node[head] at (5.25,5.15) {(b) Automatic Initialization};
\node[head] at (14.80,5.15) {(c) Iterative Joint Refinement};
\node[head] at (22.525,4.45) {Output};

\node[inner sep=0pt] (camIcon) at (.85,3.75) {\includegraphics[width=1.2cm]{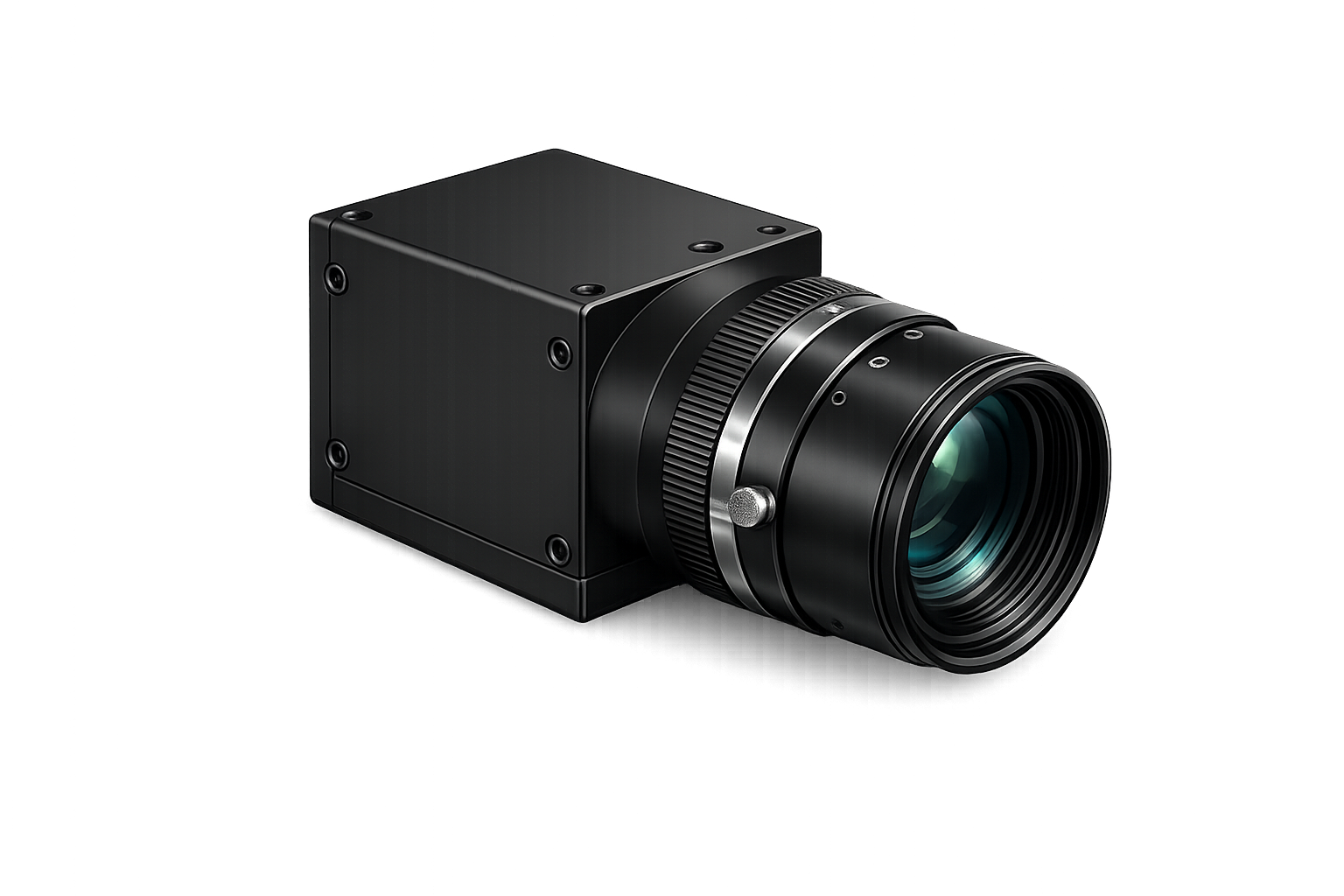}};
\node[tinylabel] at (.85,3.3) {RGB camera};

\node[inner sep=0pt] (lidarIcon) at (.85,2.15) {\includegraphics[width=.7cm]{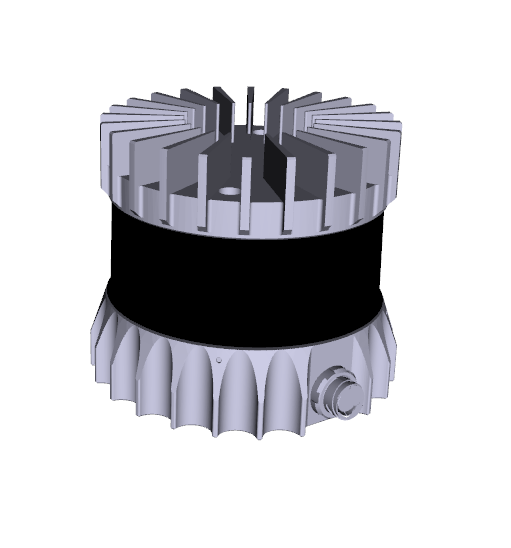}};
\node[tinylabel] at (.85,1.60) {3D LiDAR\\scanner};

\coordinate (camOut)   at (1.82,3.75);
\coordinate (lidarOut) at (1.94,2.15);

\draw[datalidarX] (lidarIcon.east) -- ($(lidarOut)+(0.4pt,0.)$);

\node[
    stage,
    minimum width=2.30cm,
    minimum height=3.35cm,
    anchor=south west
] (sfmBox) at (2.25,1.35) {};

\node[stitle,anchor=north] at ($(sfmBox.north)+(0,-.10)$)
    {Camera-parameter\\initialization\\SfM / COLMAP};

\node[inner sep=0pt] at ($(sfmBox.center)+(0,.04)$)
    {\includegraphics[width=2.2cm]{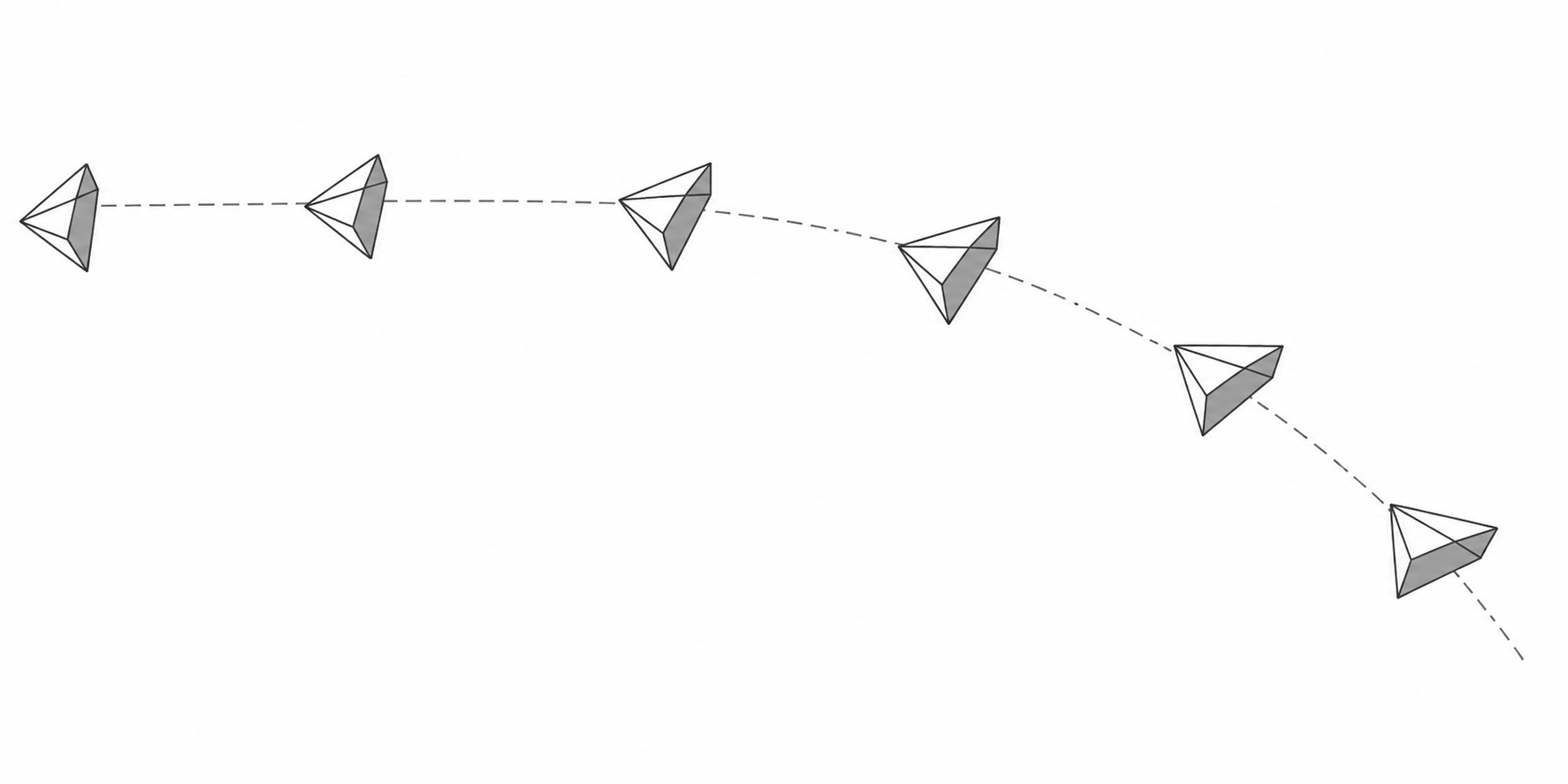}};

\node[font=\sffamily\bfseries\small] (KD0)
    at ($(sfmBox.south)+(0,.28)$)
    {$\mathbf K_0,\mathbf D_0$};

\node[
    stage,
    minimum width=2.92cm,
    minimum height=3.35cm,
    anchor=south west
] (extBox) at (5.10,1.35) {};

\node[stitle,anchor=north] at ($(extBox.north)+(0,-.10)$)
    {Extrinsic initialization\\Motion alignment~\cite{petek_automatic_2024}};

\coordinate (v0) at ($(extBox.center)+(-1.04,.25)$);
\coordinate (v1) at ($(extBox.center)+(1.04,.30)$);
\coordinate (l0) at ($(extBox.center)+(-1.04,-.55)$);
\coordinate (l1) at ($(extBox.center)+(1.04,-.50)$);

\draw[trajvis] (v0) .. controls +(.62,-.22) and +(-.62,.16) .. (v1);
\draw[trajlidar] (l0) .. controls +(.62,-.20) and +(-.62,.15) .. (l1);

\draw[densely dashed,gray!70] ($(extBox.center)+(-.66,.11)$) -- ($(extBox.center)+(-.66,-.48)$);
\draw[densely dashed,gray!70] ($(extBox.center)+(0,.09)$)    -- ($(extBox.center)+(0,-.47)$);
\draw[densely dashed,gray!70] ($(extBox.center)+(.66,.13)$)  -- ($(extBox.center)+(.66,-.43)$);

\node[tinylabel,text=teal!65!black]  at ($(extBox.center)+(0,.71)$)  {visual trajectory};
\node[tinylabel,text=green!45!black] at ($(extBox.center)+(0,-.95)$) {LiDAR trajectory};

\node[font=\sffamily\bfseries\small] (T0)
    at ($(extBox.south)+(0,.28)$)
    {$\mathbf T_0$};

\draw[datacam] (camIcon.east) -- (sfmBox.west |- camOut);

\draw[flow] (sfmBox.east) -- (extBox.west)
    node[midway,above,font=\sffamily\scriptsize]{$\mathbf K_0$}
    node[midway,below,font=\sffamily\scriptsize]{$\mathbf D_0$};

\coordinate (rgbLaneStart)   at (1.55,.85);
\coordinate (lidarLaneStart) at (1.55,.6);
\coordinate (rgbLaneEnd)     at (8.43,.85);
\coordinate (lidarLaneEnd)   at (8.55,.6);
\coordinate (rgbLaneEnd2)     at (8.85,1.2);
\coordinate (lidarLaneEnd2)   at (8.85,.95);

\draw[datacam]
    (camOut) -- (camOut |- rgbLaneStart)
    -- node[pos=.18,
            fill=white,
            inner sep=.7pt,
            font=\sffamily\scriptsize,
            text=blue!72!black]
       {RGB images}
    (rgbLaneEnd) |- (rgbLaneEnd2);

\draw[datalidar]
    (lidarOut) -- (lidarOut |- lidarLaneStart)
    -- node[pos=.18,
            fill=white,
            inner sep=.7pt,
            font=\sffamily\scriptsize,
            text=cyan!55!blue!75!black]
       {LiDAR scans}
    (lidarLaneEnd) |- (lidarLaneEnd2);

\draw[datacam](6.15,.85) -- (6.15,0 |- extBox.south);
\draw[datalidar](6.86,.6) -- (6.86,0 |- extBox.south);

\node[
    stage,
    minimum width=3.05cm,
    minimum height=3.35cm,
    anchor=south west
] (proj) at (9.18,1.35) {};

\node[
    stage,
    minimum width=3.20cm,
    minimum height=3.35cm,
    anchor=south west
] (corr) at (12.42,1.35) {};

\node[
    stage,
    minimum width=2.90cm,
    minimum height=3.35cm,
    anchor=south west
] (sel) at (15.82,1.35) {};

\node[
    stage,
    minimum width=1.55cm,
    minimum height=3.35cm,
    anchor=south west
] (opt) at (18.92,1.35) {};

\node[stitle,anchor=north] at ($(proj.north)+(0,-.10)$)
    {LiDAR projection with\\current $\mathbf K,\mathbf D,\mathbf T$};

\node[stitle,anchor=north] at ($(corr.north)+(0,-.10)$)
    {Deep pixel-point\\correspondence inference};

\node[stitle,anchor=north] at ($(sel.north)+(0,-.10)$)
    {Correspondence\\selection};

\node[stitle,anchor=north] at ($(opt.north)+(0,-.10)$)
    {Joint\\nonlinear\\optimization};

\node[inner sep=0pt] at ($(proj.center)+(0,-.1)$)
    {\includegraphics[width=2.81cm,trim=500 0 100 0,clip]{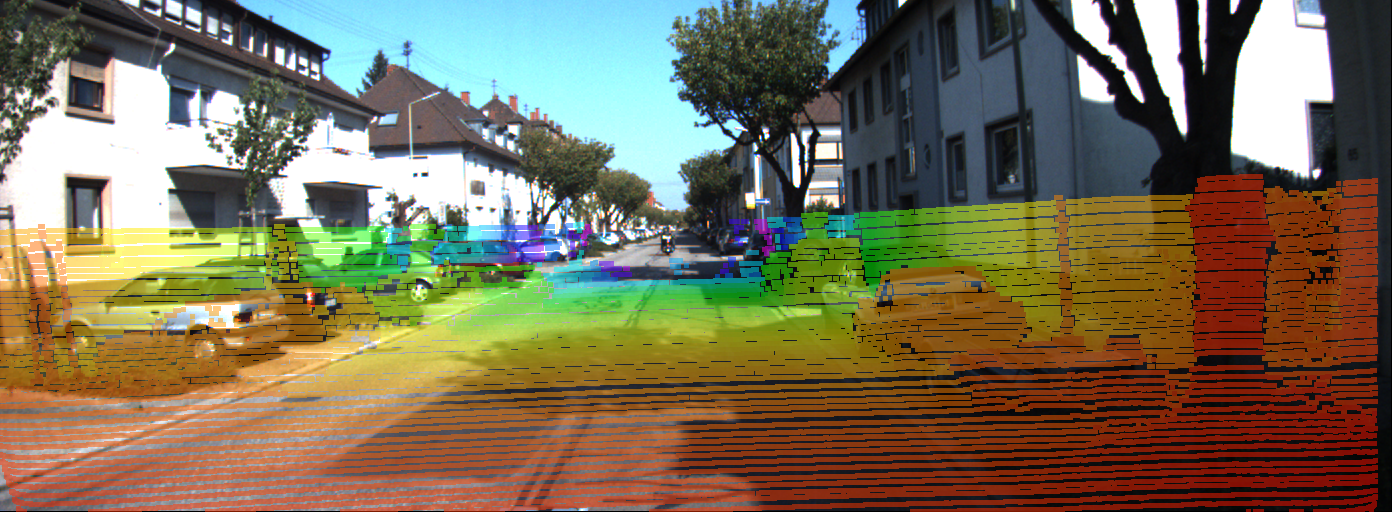}};

\node[inner sep=0pt] at ($(corr.center)+(0,-.1)$)
    {\includegraphics[width=2.96cm,trim=500 0 100 0,clip]{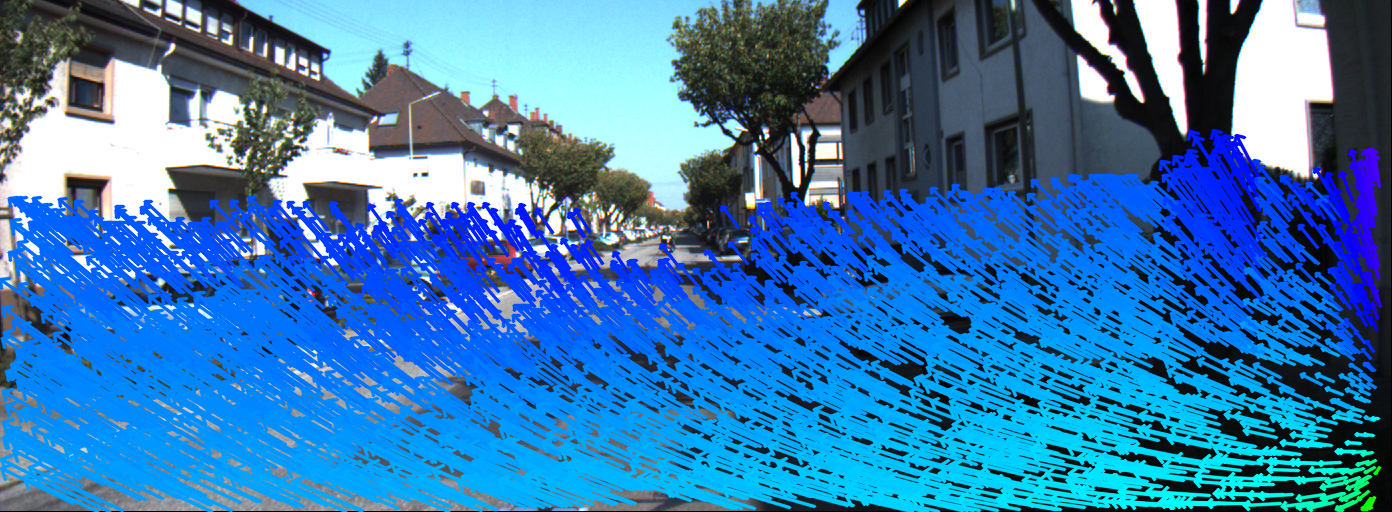}};

\node[inner sep=0pt] at ($(sel.center)+(0,-.1)$)
    {\includegraphics[width=2.66cm,trim=500 0 100 0,clip]{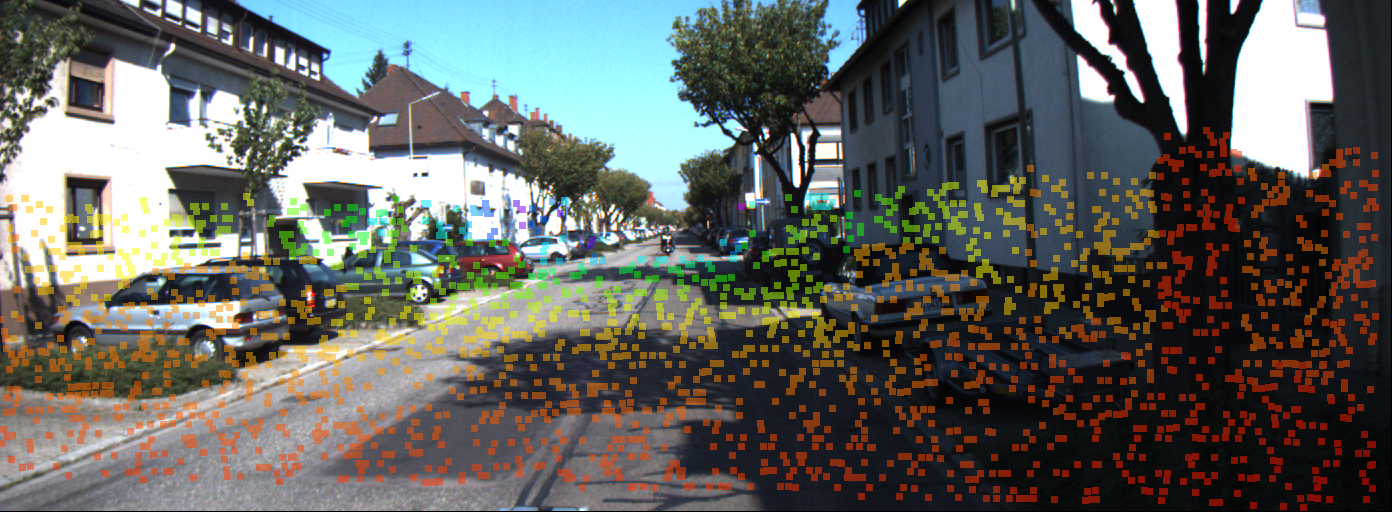}};

\node[
    font=\sffamily\scriptsize,
    align=center
] at ($(opt.center)+(0,-.3)$)
{
    $\displaystyle
    \argmin_{\boldsymbol{\theta}_{K,D},\boldsymbol{\theta}_T}$\\[4pt]
    $\displaystyle
    J_{\mathrm{corr}} +$\\[2pt]
    $\displaystyle
    \lambda_c J_{\mathrm{prior}}$
};

\draw[flow] (proj.east) -- (corr.west);
\draw[flow] (corr.east) -- (sel.west);
\draw[flow] (sel.east)  -- (opt.west);

\node[tinylabel] at ($(proj.south)+(0,.28)$)
    {current projection};

\node[tinylabel] at ($(corr.south)+(0,.28)$)
    {predicted flow field};

\node[tinylabel] at ($(sel.south)+(0,.28)$)
    {selected matches};

\draw[flow] (extBox.east) -- ++(.26,0) |- (proj.west)
    node[pos=.65,above,font=\sffamily\scriptsize,align=left]
    {$\mathbf K_0$,\\$\mathbf D_0$}
    node[pos=.65,below,font=\sffamily\scriptsize]
    {$\mathbf T_0$};

\draw[feedback]
    (opt.south) -- ++(0,-.55) -| (proj.south)
    node[pos=.43,font=\sffamily\scriptsize,fill=blue!2,inner sep=1.2pt,]
    {updated $\mathbf K,\mathbf D,\mathbf T$};

\draw[flow] (opt.east) -- (opt.east -| 21.00,2.93);

\node[font=\sffamily\bfseries\small] at (22.525,3.9)
    {$\mathbf K^*,\mathbf D^*,\mathbf T^*$};

\node[inner sep=0pt] at (22.525,2.8)
    {\includegraphics[width=2.72cm,trim=500 0 100 0,clip]{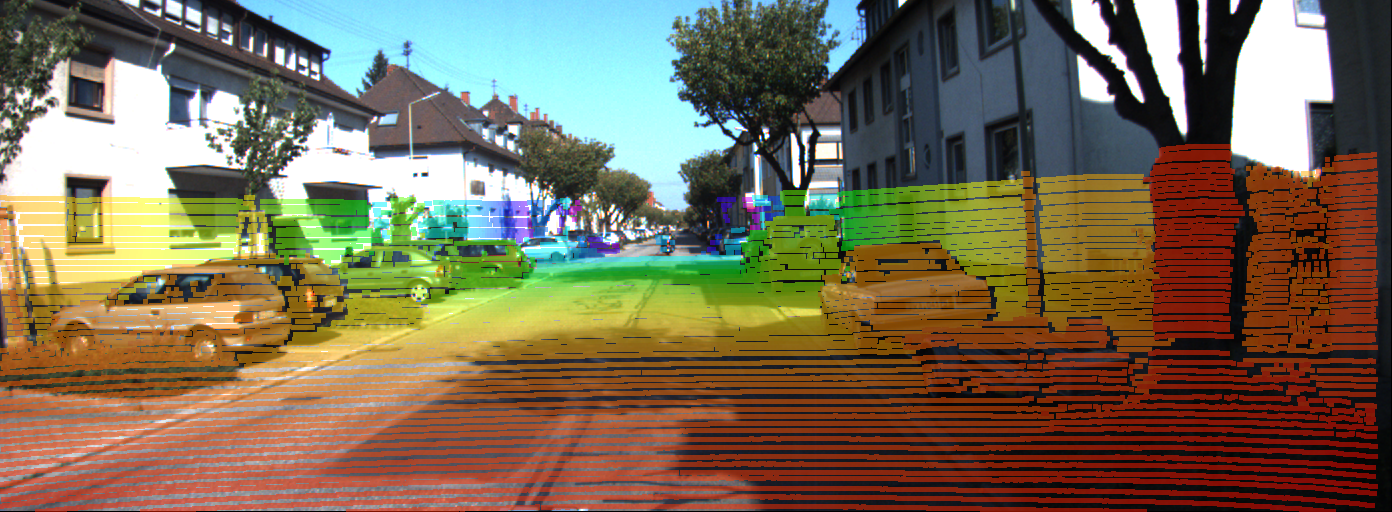}};

\node[tinylabel] at (22.525,1.7)
    {final projection};

\end{tikzpicture}%
}
\vspace{-1.em}
\caption{Overview of the proposed joint camera-LiDAR calibration framework.
Structure-from-motion (SfM) provides initial camera parameters and visual-LiDAR motion alignment initializes extrinsics.
During iterative refinement, current intrinsic, distortion, and extrinsic estimates $\mathbf K,\mathbf D,\mathbf T$ define LiDAR projection input into the correspondence network.
Predicted pixel-point correspondences are post-processed and used for joint nonlinear optimization.
Updated calibration parameters are fed back into the next iteration.
After convergence, the final parameters define the calibrated projection shown at the output.\looseness=-1}
\label{fig:pipeline}
\vspace{-.5em}
\end{figure*}
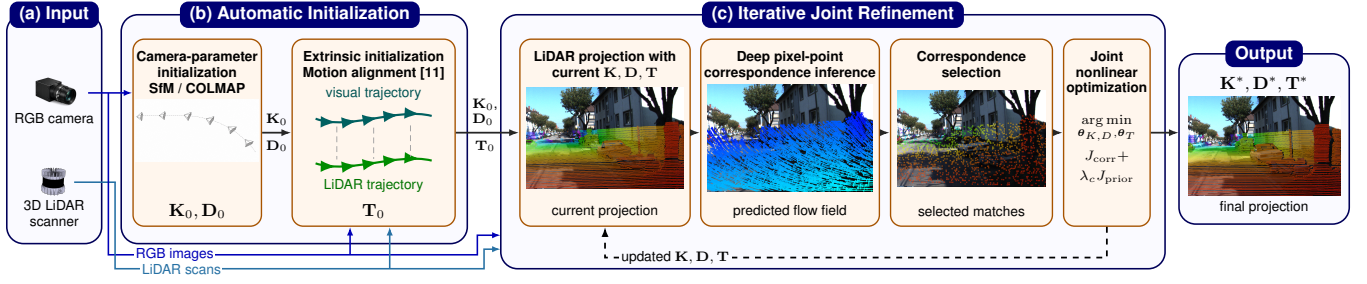

Given synchronized camera images and LiDAR scans, our framework combines camera-parameter initialization, motion-based extrinsic initialization, intrinsic- and distortion-aware correspondence prediction, and alternating nonlinear refinement of the camera parameters and relative sensor transformation.
Fig.~\ref{fig:pipeline} illustrates the entire approach.

\subsection{Camera Model and Projection}

We employ a pinhole camera model with radial-tangential distortion.
Let $\mathbf{X}\in\mathbb{R}^3$ denote a LiDAR point and
\begin{equation}
\mathbf{T}
=
\begin{bmatrix}
\mathbf{R} & \mathbf{t}\\
\mathbf{0}^{\top} & 1
\end{bmatrix}
\in\mathbb{SE}(3)
\end{equation}
be the transformation from the LiDAR to the camera frame, with
$\mathbf{R}\in\mathbb{SO}(3)$, corresponding unit quaternion $\mathbf{q}_{R}$, and $\mathbf{t}\in\mathbb{R}^3$.
The transformed point is
\begin{equation}
\mathbf{X}_{c}
=
\begin{bmatrix}
X_c & Y_c & Z_c
\end{bmatrix}^{\top}
=
\mathbf{R}\mathbf{X}+\mathbf{t}\,.
\label{eq:proj_start}
\end{equation}
For $Z_c>0$, its normalized image coordinates are
\begin{equation}
x=\frac{X_c}{Z_c},
\qquad
y=\frac{Y_c}{Z_c}\,,
\qquad
r^2 = x^2 + y^2\,.
\end{equation}
Using the distortion parameters $\mathbf{D}=[k_1,k_2,t_1,t_2,k_3]^\top$,
radial distortion is modeled as
\begin{equation}
\gamma(r)
=
1+k_1r^2+k_2r^4+k_3r^6.
\end{equation}
Together with tangential distortion, the distorted coordinates are
\begin{align}
x_d &= x \gamma(r) + 2 t_1 xy + t_2 (r^2 + 2x^2)\,, \\
y_d &= y \gamma(r) + 2 t_2 xy + t_1 (r^2 + 2y^2)\,.
\end{align}
Finally, pixel coordinates are obtained via
\begin{equation}
\mathbf{u}
=
\mathbf{K}
\begin{bmatrix}
x_d \\
y_d \\
1
\end{bmatrix},
\text{where}\;
\mathbf{K} =
\begin{bmatrix}
f_x & 0   & c_x \\
0   & f_y & c_y \\
0   & 0   & 1
\end{bmatrix}\,,
\label{eq:proj_end}
\end{equation}
with focal lengths $f_x, f_y$ and optical center $\mathbf{c} = [c_x, c_y]^\top$.
We denote the complete projection by $\Pi(\mathbf{X};\mathbf{K},\mathbf{D},\mathbf{T})$.
Our framework jointly estimates the camera parameters $\mathbf{K},\mathbf{D}$ and the relative transformation $\mathbf{T}$.

Polynomial radial distortion models may become non-monotonic outside their valid domain, causing rays far outside the field of view to fold back into the image~\cite{leotta_maximum_2022}.
For the distorted radius
\begin{equation}
r_d= r\gamma\left(r\right) = r\left(1+k_1r^2+k_2r^4+k_3r^6\right)\,,
\end{equation}
the maximum valid normalized image radius $r_{\max}$ is determined by the smallest positive root of
\begin{equation}
\frac{\partial r_d}{\partial r}
=
1+3k_1r^2+5k_2r^4+7k_3r^6
=
0\,.
\label{eq:rmax}
\end{equation}
If no positive root exists, the radial model remains monotonic for $r>0$, and we set $r_{\max}=\infty$.
For finite $r_{\max}$, we reject rays with $r>r_{\max}$ during training, image augmentation, correspondence inference, and nonlinear optimization.

\subsection{Intrinsic- and Distortion-Aware Correspondence Learning}

We employ the CMRNext architecture~\cite{cattaneo_cmrnext_2025} for pixel-point correspondence prediction. Each LiDAR scan is represented as a sparse image obtained by projecting it with the current calibration estimate, and the network predicts a dense displacement field between the resulting LiDAR pseudo-depth image and the corresponding RGB image. The displacement at each valid projected LiDAR point defines a pixel-point correspondence.

While the original CMRNext training procedure assumes rectified images with pre-calibrated intrinsics and perturbs only the camera-LiDAR extrinsics, we extend it to raw images with unknown camera parameters by additionally perturbing focal lengths, principal point, and distortion parameters when generating the projected LiDAR input.
We sample pinhole parameters using bounded relative perturbations around their reference values.
We perturb radial distortion parameters while preserving plausible signs and relative magnitudes, and sample tangential parameters in a bounded interval around zero. 
As in CMRNext, we train multiple network instances for decreasing perturbation ranges and use them sequentially during refinement.

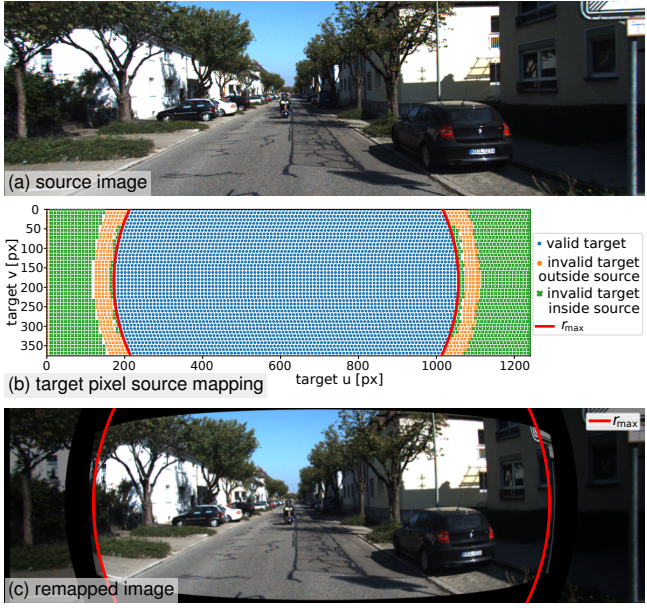
\begin{figure}[t]
    \centering
    \begin{tikzpicture}

        \node[
            anchor=south west,
            inner sep=0
        ] (p1) at (0,0) {%
            \includegraphics[
                width=0.98\columnwidth
            ]{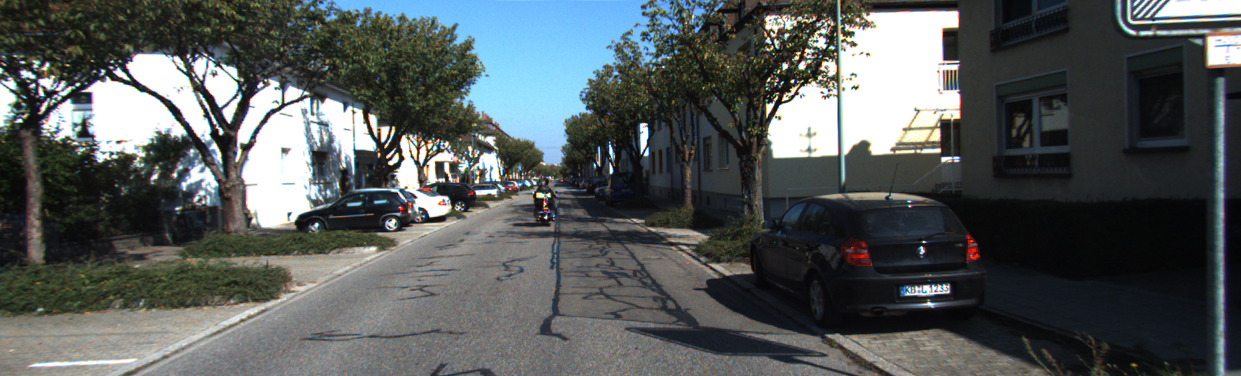}%
        };

        \node[
            anchor=south west,
            xshift=0.0cm,
            yshift=0.0cm,
            inner sep=0.12em,
            fill=gray!20,
            fill opacity=0.65,
            text opacity=1,
            font=\sffamily\scriptsize
        ] at (p1.south west) {(a) source image};

        \node[
            anchor=north west,
            inner sep=0,
            yshift=-0.10cm
        ] (p2) at (p1.south west) {%
            \includegraphics[
                width=0.98\columnwidth
            ]{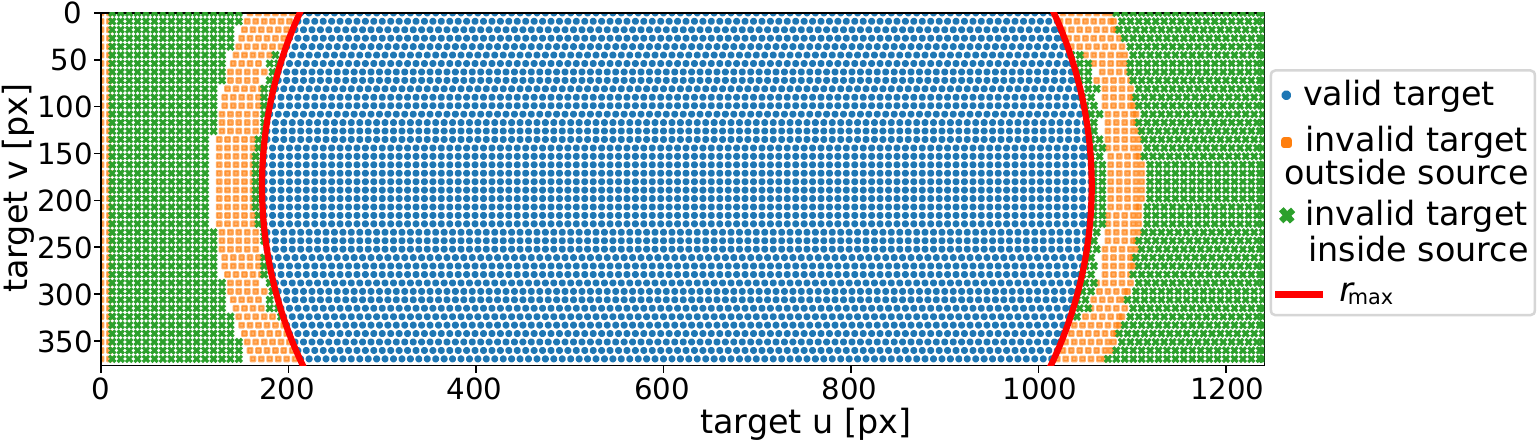}%
        };

        \node[
            anchor=south west,
            xshift=0.0cm,
            yshift=-0.12cm,
            inner sep=0.12em,
            fill=gray!20,
            fill opacity=0.65,
            text opacity=1,
            font=\sffamily\scriptsize
        ] at (p2.south west) {(b) target pixel source mapping};

        \node[
            anchor=north west,
            inner sep=0,
            yshift=-0.25cm
        ] (p3) at (p2.south west) {%
            \includegraphics[
                width=0.98\columnwidth
            ]{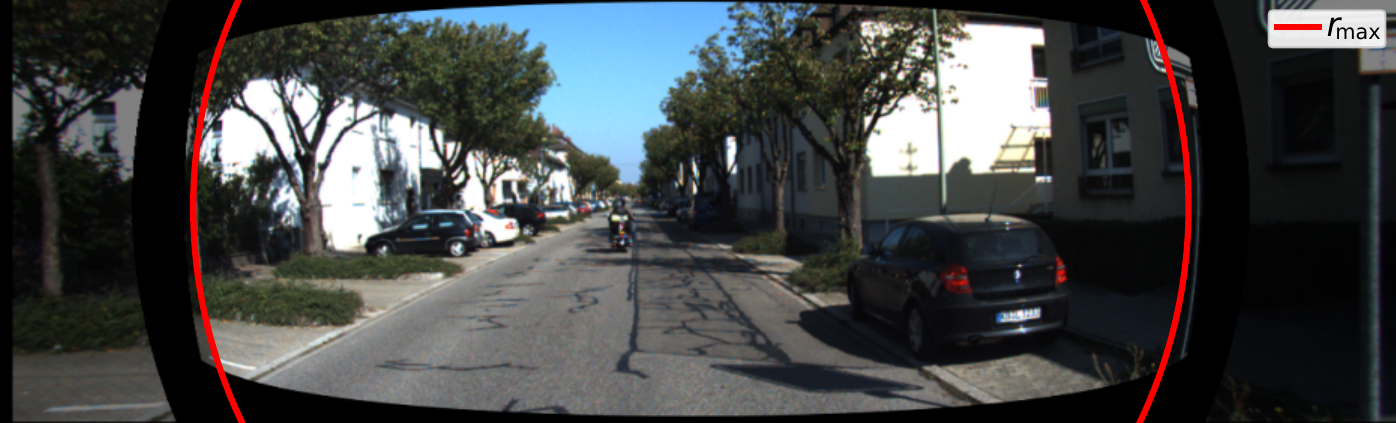}%
        };

        \node[
            anchor=south west,
            xshift=0.0cm,
            yshift=0.0cm,
            inner sep=0.12em,
            fill=gray!20,
            fill opacity=0.65,
            text opacity=1,
            font=\sffamily\scriptsize
        ] at (p3.south west) {(c) remapped image};

    \end{tikzpicture}
	\vspace{-.5em}
    \caption{
        Synthetic camera-parameter augmentation.
        (a): source image.
        (b): target pixels colored by geometric validity and source-image mapping.
        (c): remapped image with invalid pixels darkened; only valid pixels are retained.
    }
    \label{fig:distortion_augmentation}
    \vspace{-.5em}
\end{figure}

Parameter perturbation alone changes the LiDAR projection but does not expose the network to previously unseen distorted cameras.
We therefore synthesize images under sampled camera parameters as illustrated in Fig.~\ref{fig:distortion_augmentation}.
Given a source camera
$\mathcal{C}_s=(\mathbf{K}_s,\mathbf{D}_s)$
and a sampled target camera
$\mathcal{C}_t=(\mathbf{K}_t,\mathbf{D}_t)$,
each target pixel $\mathbf{u}_t$ is mapped to a source pixel by
\begin{equation}
\mathbf{u}_s
=
\Pi\!\left(
\Pi^{-1}(\mathbf{u}_t;\mathbf{K}_t,\mathbf{D}_t);
\mathbf{K}_s,\mathbf{D}_s
\right),
\label{eq:camera_remap}
\end{equation}
where $\Pi^{-1}$ maps the target pixel to its normalized camera ray and $\Pi$ reprojects that ray into the source image.
Evaluating Eq.~\eqref{eq:camera_remap} over the complete target image yields dense backward-warp fields suitable for standard image resampling.
However, target pixels with $r>r_{\max}$ lie outside the valid distortion domain and either map outside the source image or fold back onto source-image coordinates, potentially duplicating image content (cf. Fig.~\ref{fig:distortion_augmentation}\,(b),\,(c)).
Therefore, a target pixel is retained only if its normalized ray lies within the valid distortion domain, satisfying $r\leq r_{\max}$.

This combination of parameter perturbation and image-domain remapping trains the correspondence network to handle both projection errors and appearance changes induced by previously unseen pinhole and distortion parameters. At inference time, the current calibration estimate is used to project the LiDAR scan into the raw image, and the network predicts the correspondence set $\tilde{\mathcal{M}}$ from the resulting projection and the raw image. This set is post-processed into $\mathcal{M}$ for nonlinear optimization, as detailed below.

\subsection{Joint Nonlinear Optimization}
\label{sec:method_opt}

Let
$\mathcal{M}=\{(\mathbf{X}_{i},\hat{\mathbf{u}}_{i})\}$
denote the correspondences used for optimization, where
$\hat{\mathbf{u}}_{i}$ is the image location predicted for LiDAR point $\mathbf{X}_{i}$.
We jointly estimate the camera parameters and relative sensor transformation
\begin{equation}
\boldsymbol{\theta}_{K,D}
=
(f_x,f_y,c_x,c_y,\mathbf{D}),
\qquad
\boldsymbol{\theta}_{T}
=
(\mathbf{q}_{R},\mathbf{t})
\end{equation}
by minimizing the robust reprojection error with an additional weak prior on the principal point:
\begin{align}
\boldsymbol{\theta}^{*}
&=
\argmin_{\boldsymbol{\theta}_{K,D},\boldsymbol{\theta}_{T}}
\left(
J_{\mathrm{corr}}
+
\lambda_c J_{\mathrm{prior}}
\right),
\label{eq:joint_optimization}
\\
J_{\mathrm{corr}}
&=
\sum_i
\rho\!\left(
\left\|
\hat{\mathbf{u}}_i
-
\Pi(
\mathbf{X}_i;
\boldsymbol{\theta}_{K,D},
\boldsymbol{\theta}_{T}
)
\right\|_2^2
\right),
\label{eq:joint_corr}
\\
J_{\mathrm{prior}}
&=
\left\|
\mathbf{c}-\mathbf{c}_0
\right\|_2^2 .
\label{eq:joint_prior}
\end{align}
Here, $\rho(\cdot)$ is the Cauchy loss used by MDPCalib~\cite{petek_automatic_2024}, $\mathbf{c}_0$ is the principal-point initialization obtained from SfM, and $\lambda_c$ controls the strength of the prior.
As the principal point~$\mathbf{c}$ is weakly constrained by the pixel-point correspondences, $J_{\mathrm{prior}}$ is introduced to stabilize the optimization while still allowing it to deviate from the initialization; its benefit is evaluated in Tab.~\ref{tab:kitti_ablation}.
We solve Eq.~\eqref{eq:joint_optimization} using Ceres nonlinear least squares~\cite{Agarwal_Ceres_Solver_2022}, parameterizing rotation on the unit-quaternion manifold.

To obtain $\mathcal{M}$ from $\tilde{\mathcal{M}}$, we first discard correspondences whose projections are invalid under the current camera parameters, i.e., outside the image or beyond the distortion validity radius.
We then process the remaining correspondences through RANSAC-based inlier selection and uniform subsampling.
Specifically, we undistort the predicted image endpoints using the current camera parameters and apply PnP-RANSAC to obtain correspondence inliers, which are uniformly subsampled to $10\%$ of the original correspondence set.
The first iteration omits inlier selection because the initial camera parameters may not yet support a consistent rigid transformation estimate.

\subsection{Alternating Correspondence and Calibration Refinement}

Existing correspondence-based methods use PnP or a single nonlinear optimization to update only the extrinsics while keeping the camera parameters fixed.
In our framework, each correspondence stage is instead followed by the joint optimization in Eq.~\eqref{eq:joint_optimization}.
Updated intrinsic and extrinsic parameters change the LiDAR projection used by the subsequent network stage, coupling correspondence prediction and calibration refinement, as detailed in Alg.~\ref{alg:joint_calibration}. We refine camera parameters for the first
$N_{\mathrm{intr}}=5$ iterations and subsequently keep them fixed while continuing extrinsic refinement until
$N_{\mathrm{total}}=11$.
Freezing the camera parameters early prevents residual correspondence noise from causing drift in weakly observable components while allowing the extrinsics to continue improving.
We select the iteration counts through hyperparameter evaluation (cf. Fig.~\ref{fig:iteration_study}).

\begin{figure}[t]
\vspace{-1.4em}
\begin{algorithm}[H]
\caption{Joint Camera-LiDAR Calibration}
\label{alg:joint_calibration}
\begin{algorithmic}[1]
\Require Synced images and LiDAR scans,
initial $\boldsymbol{\theta}_{K,D}^{0}$,
$\boldsymbol{\theta}_{T}^{0}$
\Ensure Refined calibration
\For{$j=1,\ldots,N_{\mathrm{total}}$}
\State Select correspondence model for iteration $j$
\State Project LiDAR points using
$\boldsymbol{\theta}_{K,D}^{j-1},\boldsymbol{\theta}_{T}^{j-1}$
\State Predict pixel-point correspondences $\tilde{\mathcal{M}}^{j}$
\State Post-process and subsample $\tilde{\mathcal{M}}^{j}$ to $\mathcal{M}^{j}$
\If{$j\leq N_{\mathrm{intr}}$}
\State Jointly optimize
$\boldsymbol{\theta}_{K,D}^{j},\boldsymbol{\theta}_{T}^{j}$
using Eq.~\eqref{eq:joint_optimization} and $\mathcal{M}^{j}$\looseness=-1
\Else
\State Set
$\boldsymbol{\theta}_{K,D}^{j}
\gets
\boldsymbol{\theta}_{K,D}^{N_{\mathrm{intr}}}$
and optimize only $\boldsymbol{\theta}_{T}^{j}$
\EndIf
\EndFor
\State \Return
$\boldsymbol{\theta}_{K,D}^{N_{\mathrm{intr}}},
\boldsymbol{\theta}_{T}^{N_{\mathrm{total}}}$
\end{algorithmic}
\vspace{-.2em}
\end{algorithm}
\vspace{-2.2em}
\end{figure}

\subsection{Initialization}
\label{sec:method_init}

The alternating refinement procedure requires initial estimates that yield sufficient overlap between the projected LiDAR scan and the camera image.
We initialize the camera parameters through structure-from-motion (SfM) and the rigid sensor transformation through motion alignment, not requiring any manual initial guess. To obtain initial camera parameters without a calibration target, we employ COLMAP~\cite{schoenberger2016sfm} on a subset of $N_{\mathrm{init}}$ images exhibiting sufficient translational and rotational motion. The reconstruction uses COLMAP's OPENCV camera model, estimating focal lengths, principal point, and radial-tangential distortion with four parameters. We initialize the additional coefficient $k_3=0$. For extrinsic initialization, we follow the motion-alignment of MDPCalib~\cite{petek_automatic_2024}. ORB-SLAM3~\cite{ORBSLAM3_TRO} provides the visual trajectory, while FAST-LIO2~\cite{xu_fastlio2_2022}, operated without IMU measurements, provides the LiDAR trajectory.
Unlike MDPCalib, which uses ground-truth camera intrinsics, we initialize ORB-SLAM3 with the COLMAP estimates, preserving a fully practical pipeline without privileged calibration information while still obtaining sufficiently accurate trajectory alignment.

\section{Experimental Evaluation}
\label{sec:experiments}

We evaluate the proposed method on the public KITTI dataset and three real robotic platforms.
The experiments assess three aspects: generalization to an unseen camera-LiDAR configuration, the accuracy of joint intrinsic-extrinsic calibration, and the contribution of the proposed training and optimization components.

\subsection{Experimental Setup}

Our training setup follows the protocol of MDPCalib~\cite{petek_automatic_2024} and CMRNext~\cite{cattaneo_cmrnext_2025}.
We train on the rectified left camera of KITTI~\cite{kitti_geiger_2012} sequences 01--21, together with one rectified camera from Argoverse~\cite{argoverse_chang_2019} and PandaSet~\cite{pandaset_xiao_2021}, respectively.
To calibrate non-rectified cameras, we also include raw images where available (i.e., the KITTI left camera in sequences 01, 02, and 04-10) and apply the proposed camera-parameter augmentation during training.
We evaluate on the raw right camera of KITTI sequence 00, which differs from the training data in both sequence and camera-LiDAR configuration, to assess generalization to an unseen sensor setup. %

\begin{table}[t]
\centering
\caption{Perturbation ranges during network training.
}
\vspace{-1em}
\label{tab:network_schedule}
\setlength{\tabcolsep}{4pt}
\renewcommand{\arraystretch}{1.05}
\begin{tabular}{lccccc}
\toprule
Model & Transl. & Rot. & Intr. & Dist. & Used in iter. \\
\midrule
Coarse & $\phantom{0} 1.0\,\mathrm{m}$ & $5.0^\circ$ & $10\%^{\dagger}$ & $15\%$ & $1$ \\
Medium & $\phantom{0} 0.1\,\mathrm{m}$ & $1.0^\circ$ & $2\%$ & $10\%$ & $2$--$3$ \\
Fine & $0.05\,\mathrm{m}$ & $0.5^\circ$ & $1\%$ & $5\%$ & $4+$ \\
\bottomrule
\end{tabular}

\vspace{0.3em}
\footnotesize
\raggedright
$^{\dagger}$ coarse: principal point additionally sampled at image center ($p=0.3$).
\vspace{-1.em}
\end{table}

At inference time, we use three correspondence networks trained with decreasing perturbation ranges, denoted \textit{coarse}, \textit{medium}, and \textit{fine}, as summarized in Tab.~\ref{tab:network_schedule}.
We use the coarse model for the first refinement iteration, the medium model for the next two iterations, and the fine model thereafter.
Unless stated otherwise, camera parameters are refined during the first $N_{\mathrm{intr}}=5$ iterations, after which they are fixed while extrinsic refinement continues until $N_{\mathrm{total}}=11$, as selected by the hyperparameter study in Sec.~\ref{sec:eval_ablation}. 
Each calibration run uses 100 synchronized image-scan pairs, following~\cite{petek_automatic_2024}. 
For each image-scan pair, the network predicts dense pixel-point correspondences, from which we retain $10\%$ using uniform image-space subsampling after the RANSAC inlier selection stage described in Sec.~\ref{sec:method_opt}. 
We initialize camera parameters from ground truth (GT), COLMAP SfM (COL), or perturbed versions (GT$^+$, COL$^+$).
For perturbed initialization, we add independent uniform noise of $\pm 10\%$ to the initial camera parameters and set the principal point to the image center.
We initialize extrinsics using the motion-alignment stage described in Sec.~\ref{sec:method_init}.

\subsection{Metrics}

We measure extrinsic calibration accuracy by translation and rotation errors w.r.t. the reference calibration.
We report component-wise translation errors along $x$, $y$, and $z$, their mean absolute value, component-wise rotation errors in roll, pitch, and yaw, and the shortest-arc rotation error. 
We evaluate intrinsic accuracy both parameter-wise and geometrically.
For focal lengths $f_x$, $f_y$, and principal point $c_x$, $c_y$, we report relative errors.
Since camera parameters have heterogeneous scales and coupled effects on the projection, we additionally report reprojection error.
Given normalized camera rays sampled uniformly over the image, we compare their projections under the estimated and reference camera parameters and report the median pixel error.
We also report a distortion-only reprojection error, where we fix the pinhole parameters to the reference values and evaluate only the estimated distortion model.
This separates distortion errors from errors in focal length and principal point. 
All results are averaged over 100 runs and reported as mean and standard deviation unless stated otherwise.

\begin{table*}[t]
\centering
\caption{Joint intrinsic and extrinsic calibration results on KITTI (Seq. 00, right camera, unseen setup).
Mean $\pm$ std over 100 runs.}\vspace{-1em}
\label{tab:kitti_main}
\setlength{\tabcolsep}{2.5pt}
\renewcommand{\arraystretch}{1.05}
\begin{tabular}{llcccccccccccccc}
\toprule
\multirow{2}{*}{Method}
&
\multirow{2}{*}{Setting}
&
\multicolumn{4}{c}{Translation [cm]}
&
\multicolumn{4}{c}{Rotation [$^\circ$]}
&
\multicolumn{4}{c}{Intrinsics [\%]}
&
\multicolumn{2}{c}{Reproj. [px]}
\\
\cmidrule(lr){3-6}
\cmidrule(lr){7-10}
\cmidrule(lr){11-14}
\cmidrule(lr){15-16}
&
&
$x$
&
$y$
&
$z$
&
magnitude
&
roll
&
pitch
&
yaw
&
magnitude
&
$f_x$
&
$f_y$
&
$c_x$
&
$c_y$
&
full
&
dist.
\\
\midrule
CMRNext~\cite{cattaneo_cmrnext_2025}
&
fixed GT
&
0.37 & 6.03 & 0.12 & $6.04\pm0.38$ %
&
0.01 & 0.01 & 0.28 & $0.28\pm0.006$ %
&
-- & -- & -- & --
&
-- & --
\\
MDPCalib~\cite{petek_automatic_2024}
&
fixed GT
&
0.46 & 2.84 & 0.44 & $2.92\pm0.29$ %
&
0.02 & 0.05 & 0.13 & $0.14\pm0.010$ %
&
-- & -- & -- & --
&
-- & --
\\
Ours
&
fixed GT
&
1.24 & 1.28 & 1.37 & $\mathbf{2.25}\pm\mathbf{0.05}$
&
0.02 & 0.04 & 0.02 & $\mathbf{0.05}\pm\mathbf{0.002}$
&
-- & -- & -- & --
&
-- & --
\\
\midrule
MDPCalib$^*$
&
fixed COL
&
30.2 & 5.14 & 6.18 & $31.2\pm0.40$
&
0.13 & 0.12 & 0.06 & $0.19\pm0.01$
&
2.11 & 0.26 & 0.46 & 0.25
&
5.45 & 0.86
\\
Ours
&
fixed COL
&
2.81 & 0.49 & 0.13 & $\mathbf{2.86}\pm\mathbf{0.56}$
&
0.08 & 0.09 & 0.05 & $\mathbf{0.13}\pm\mathbf{0.02}$
&
2.11 & 0.26 & 0.46 & 0.25
&
5.45 & 0.86
\\
\midrule
Ours
&
opt. GT$^+$
&
0.26 & 2.08 & 0.86 & $\mathbf{2.27}\pm\mathbf{0.09}$
&
0.02 & 0.09 & 0.05 & $\mathbf{0.11}\pm\mathbf{0.02}$
&
0.40 & 0.56 & 0.07 & 0.30
&
$\mathbf{1.14}\pm\mathbf{0.24}$ & $\mathbf{0.43}\pm\mathbf{0.06}$
\\
Ours
&
opt. COL
&
0.17 & 2.10 & 0.90 & $2.29\pm0.07$
&
0.02 & 0.09 & 0.09 & $0.13\pm0.02$
&
0.44 & 0.60 & 0.13 & 0.30
&
$1.31\pm0.32$ & $0.49\pm0.05$
\\
Ours
&
opt. COL$^+$
&
0.39 & 2.14 & 0.82 & $2.34\pm0.12$
&
0.02 & 0.08 & 0.06 & $0.11\pm0.03$
&
0.46 & 0.62 & 0.10 & 0.27
&
$1.25\pm0.38$ & $0.55\pm0.10$
\\
\bottomrule
\end{tabular}

\vspace{0.3em}
\footnotesize
\raggedright
GT: ground-truth intr. init.
COL: COLMAP intr. init.
GT$^+$ and COL$^+$: additional \(\pm10\%\) uniform perturbation.
fixed: intr. fixed during calibration.
opt.: intr. jointly optimized with extrinsics.
dist.: distortion-only reproj. error.
MDPCalib$^*$: MDPCalib applied to images undistorted using fixed COLMAP intrinsics.
Translation components follow LiDAR convention ($x$ forward, $y$ left, $z$ up).
Standard deviation reported for magnitude and reproj. only due to space.
\end{table*}

\begin{table}[t]
\centering
\caption{Ablation study on KITTI Seq. 00 (raw right camera, unseen setup; perturbed GT$^+$ intr. init.).
Mean $\pm$ std over 100 runs.}\vspace{-1em}
\label{tab:kitti_ablation}
\setlength{\tabcolsep}{2.5pt}
\renewcommand{\arraystretch}{1.05}
\begin{tabular}{ccc|ccc}
\toprule
Augm.
&
RANSAC
&
PP prior
&
Trans. [cm]
&
Rot. [$^\circ$]
&
Reproj. [px]
\\
\midrule
--
&
--
&
--
&
$5.30\pm1.48$ 
&
$2.73\pm0.09$ 
&
$42.7\pm1.95$ 
\\
\checkmark
&
--
&
--
&
$2.54\pm0.13$
&
$0.41\pm0.23$
&
$6.10\pm3.47$
\\
\checkmark
&
\checkmark
&
--
&
$2.48\pm0.24$
&
$0.24\pm0.12$
&
$4.11\pm1.84$
\\
\checkmark
&
--
&
\checkmark
&
$2.32\pm0.44$ 
&
$0.27\pm0.04$
&
$3.64\pm0.75$
\\
\checkmark
&
\checkmark
&
\checkmark
&
$\mathbf{2.27}\pm\mathbf{0.09}$
&
$\mathbf{0.11}\pm\mathbf{0.02}$
&
$\mathbf{1.14}\pm\mathbf{0.24}$
\\
\bottomrule
\end{tabular}

\vspace{0.3em}
\footnotesize
\raggedright
Augm.: training with camera-parameter augmentation.
RANSAC: inlier selection.
PP prior: principal-point prior $J_{\mathrm{prior}}$.%
\vspace{-.5em}
\end{table}

\subsection{Joint Calibration on an Unseen Camera-LiDAR Setup}

Tab.~\ref{tab:kitti_main} evaluates calibration on the unseen KITTI right-camera setup under three increasingly challenging settings: known ground-truth intrinsics, fixed automatically estimated intrinsics, and joint intrinsic-extrinsic optimization. We report translation components in LiDAR coordinate convention, with $x$ forward, $y$ left, and $z$ up. With fixed ground-truth intrinsics, our method achieves the lowest overall translation and rotation errors, reducing the translation magnitude from $2.92\,\mathrm{cm}$ for MDPCalib to $2.25\,\mathrm{cm}$ and the rotation magnitude from $0.14\,\degree$ to $0.05\,\degree$, while also reducing the variance. The baselines are dominated by lateral translation and yaw rotation errors, whereas ours is more balanced across translation components and remains small across all rotation axes. The importance of robustness to imperfect intrinsics becomes apparent when replacing the ground-truth camera parameters with the COLMAP estimate. MDPCalib$^*$, which processes images undistorted using the fixed COLMAP intrinsics, exhibits a large forward ($x$) translation error of $30.2\,\mathrm{cm}$. This behavior is consistent with sensitivity to focal-length mismatch, since an incorrect focal length primarily affects the inferred depth scale. In contrast, our method retains the previous order of magnitude, with $2.86\,\mathrm{cm}$ translation and $0.13\,\degree$ rotation error.

Finally, the joint calibration results show that optimizing intrinsics does not compromise extrinsic accuracy. Despite solving the more difficult intrinsic-extrinsic calibration problem, our method attains translation and rotation errors lower than baseline methods that assume known intrinsics, while additionally recovering focal length, principal point, and distortion. Starting from perturbed ground-truth intrinsics yields $2.27\,\mathrm{cm}$ and $0.11\,\degree$, while COLMAP initialization gives similarly stable results. %
These results demonstrate practical joint calibration without privileged intrinsic initialization.

\subsection{Ablation Study}
\label{sec:eval_ablation}
Tab.~\ref{tab:kitti_ablation} shows that the proposed components contribute to calibration accuracy and robustness.
Camera-parameter augmentation provides the largest improvement, reducing mean translation error by $2.76\,\mathrm{cm}$, rotation error by $2.32\,\degree$, and reprojection error by $36.6\,\mathrm{px}$.
RANSAC-based inlier selection mainly improves rotation accuracy, lowering the error from $0.41\,\degree$ to $0.24\,\degree$.
The principal-point prior $J_{\mathrm{prior}}$ improves both translation and rotation while substantially stabilizing the intrinsic estimate, reducing reprojection error from $6.10\,\mathrm{px}$ to $3.64\,\mathrm{px}$ and also reducing its variance.
This reflects the coupling between principal point and pose: principal-point errors can be partially compensated by changes in camera rotation and translation, making the two parameter sets difficult to separate. %
$J_{\mathrm{prior}}$ limits this ambiguity, while inlier selection suppresses geometrically inconsistent correspondences.
Combining both yields the lowest translation, rotation, and reprojection errors.

\begin{figure}[t]
\centering

\begin{tikzpicture}

\node[anchor=north west,inner sep=0] (p1) at (0,0)
{\includegraphics[width=0.97\linewidth]
{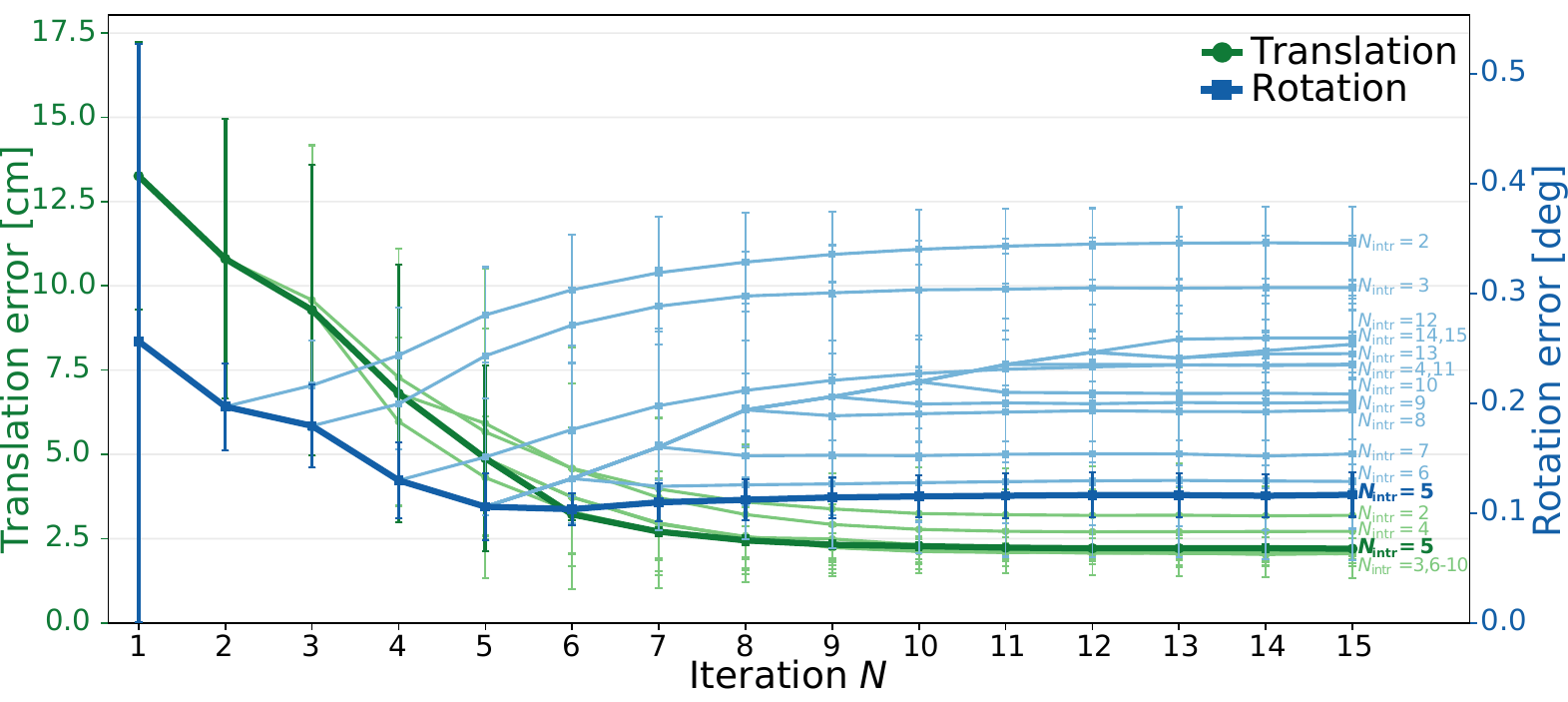}};

\node[
   anchor=north,
   yshift=-0.2cm,
   inner sep=0.em,
   fill=white,
   font=\sffamily\scriptsize
] at (p1.north) {(a) Translation and rotation};

\node[anchor=north west,inner sep=0,yshift=-0.em] (p2) at (p1.south west)
{\includegraphics[width=0.97\linewidth]
{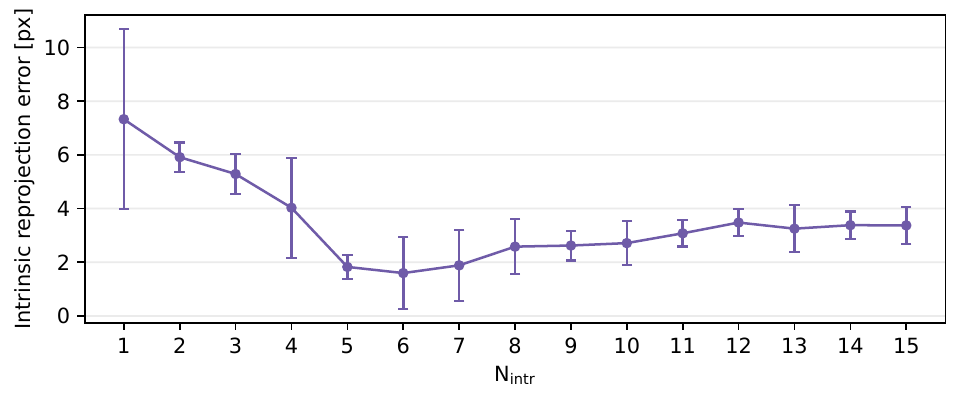}};

\node[
   anchor=north,
   yshift=-0.25cm,
   inner sep=0.em,
   fill=white,
   font=\sffamily\scriptsize
] at (p2.north) {(b) Intrinsic reprojection error};

\end{tikzpicture}
\vspace{-.7em}

\caption{
Hyperparameter study on optimization schedule:
(a) Translation and rotation errors for intrinsic refinement durations $N_{\mathrm{intr}}$ over total iterations $N$.
For $N\leq N_{\mathrm{intr}}$, each point aggregates all runs with active intrinsic optimization. For $N> N_{\mathrm{intr}}$, individual trajectories are shown for intrinsics fixed at resp. $N_{\mathrm{intr}}$.
(b) Intrinsic reprojection error over $N_{\mathrm{intr}}$.
The selected schedule $N_{\mathrm{intr}}=5$ and $N_{\mathrm{total}}=11$ provides the best overall trade-off between accurate intrinsic estimation and stable extrinsic refinement.
}
\label{fig:iteration_study}
\vspace{-.5em}
\end{figure}

We determine the optimization schedule through a hyperparameter study on KITTI, shown in Fig.~\ref{fig:iteration_study}.
The rotation error exhibits a clear optimum at $N_{\mathrm{intr}}=5$, while the translation error is largely insensitive to $N_{\mathrm{intr}}$ (Fig.~\ref{fig:iteration_study}~(a)). It mainly benefits from additional extrinsic-only refinement, with diminishing returns for $N>11$. Fig.~\ref{fig:iteration_study}~(b) confirms that refining intrinsics for at least five iterations is necessary to obtain accurate camera parameters. While $N_{\mathrm{intr}}=6$ yields a slightly lower reprojection error, it also exhibits noticeably higher variance and degrades the rotation estimate. We therefore select $N_{\mathrm{intr}}=5$ and $N_{\mathrm{total}}=11$ for all remaining experiments as the most stable overall operating point.\looseness=-1

\subsection{Real-Robot Experiments}

\begin{figure*}[t]
    \centering
    
	\colorlet{nearcolor}{cyan!85!black}
	
    \begin{tikzpicture}[
	    img/.style={inner sep=0pt, outer sep=0pt},
	    label/.style={
	        anchor=south west,
	        inner sep=0.12em,
	        fill=gray!20,
	        fill opacity=0.65,
	        text opacity=1,
	        font=\sffamily\scriptsize
	    },
	    nearhighlight/.style={draw=nearcolor, line width=1.6pt},
	    farhighlight/.style={draw=white, line width=1.2pt},
	    distlabel/.style={font=\sffamily\scriptsize\bfseries,inner sep=1pt,text=white},
	    distlabelX/.style={font=\sffamily\scriptsize\bfseries,text=white,inner sep=1pt,fill=black,fill opacity=0.38,text opacity=1,rounded corners=1pt},
	    distarrowX/.style={-{Latex[length=2mm,width=1.4mm]},line width=1.2pt,draw=white,preaction={draw=black, opacity=0.35, line width=2.8pt}},
	    spy using outlines={circle,magnification=3,size=2.1cm,every spy on node/.style={draw=white,line width=1.2pt},every spy in node/.style={draw=white,line width=1.2pt}}
	]

    \newlength{\qualimgw}
    \setlength{\qualimgw}{0.49\linewidth}
    \def\colgap{0.01\linewidth}
    \def\rowgap{0.1cm}

    \node[img, anchor=north west] (r1c1) at (0,0) {\includegraphics[width=\qualimgw,trim=0 0 0 0,clip]{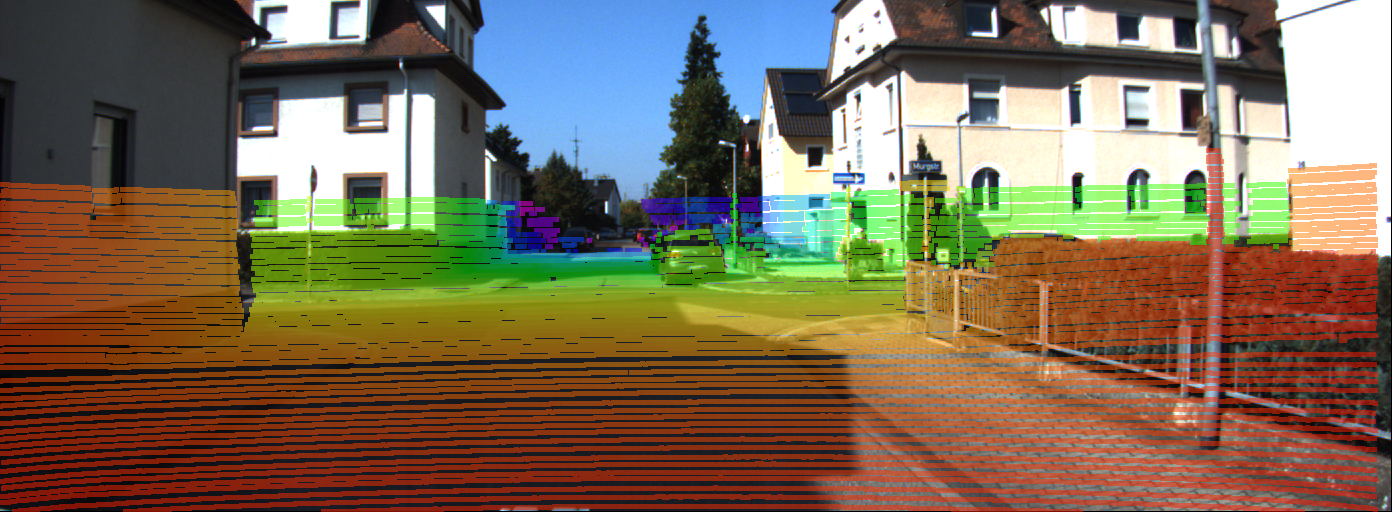}};
    \node[img, anchor=north west] (r1c2) at ($(r1c1.north east)+(\colgap,0)$) {\includegraphics[width=\qualimgw,trim=0 353px 0 140px,clip]{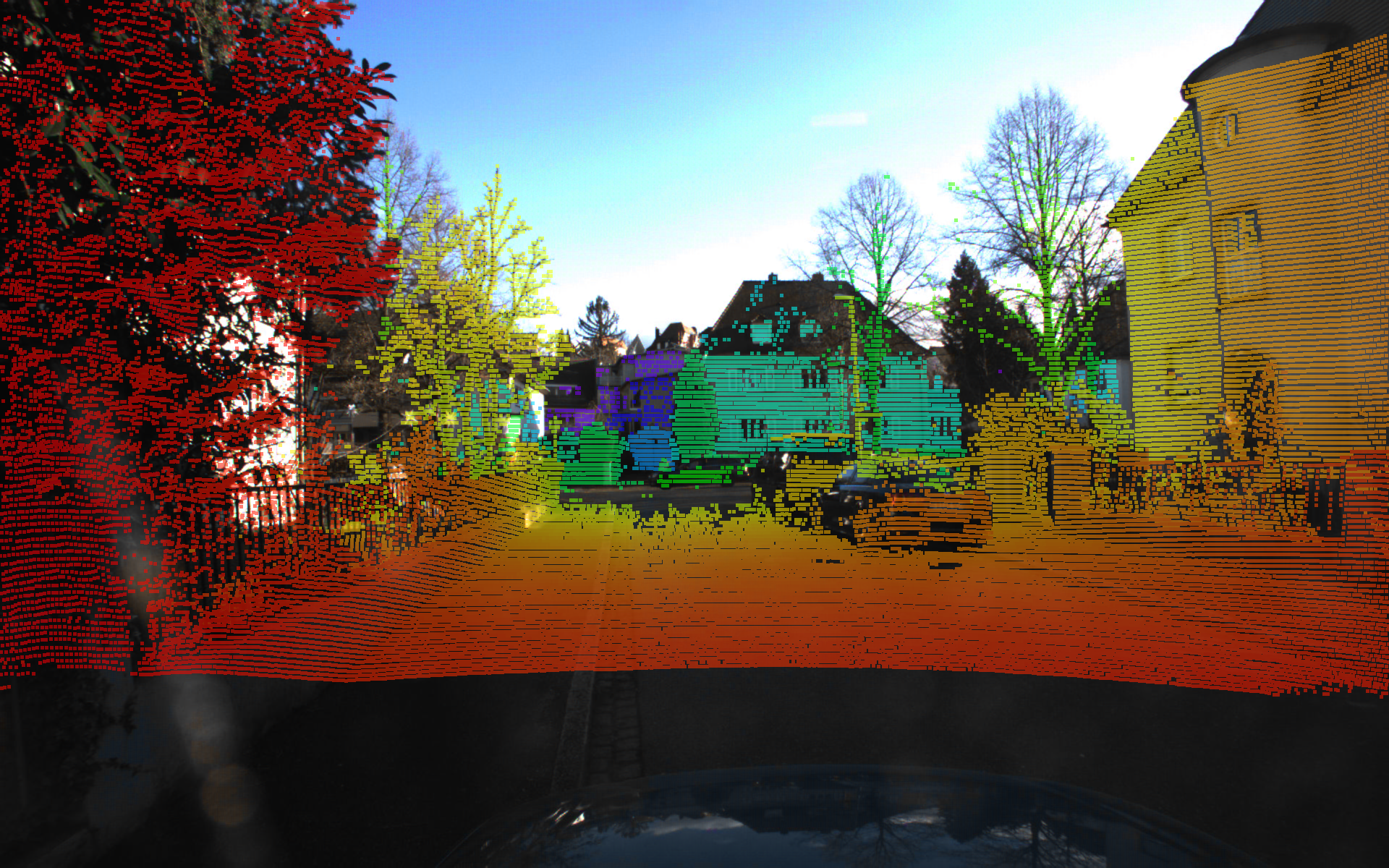}};

	\relcoord{r1c1}{kittiNear}{0.872}{0.42}
	\draw[nearhighlight] (kittiNear) ellipse[x radius=0.25cm,y radius=1.20cm];
	
	\relcoord{r1c1}{kittiFar}{0.53}{0.55}
	\relcoord{r1c1}{kittiZoomPos}{0.35}{0.37}
	\zoomspy{kittiFar}{kittiZoomPos}{3.3}{2.1cm}
	
	\relcoord{r1c2}{carFar1}{0.63}{0.6}
	\relcoord{r1c2}{carZoomPos1}{0.85}{0.45}
	\zoomspy{carFar1}{carZoomPos1}{3.3}{2.1cm}
	
	\relcoord{r1c2}{carFar2}{0.48}{0.48}
	\relcoord{r1c2}{carZoomPos2}{0.15}{0.55}
	\zoomspy{carFar2}{carZoomPos2}{3.3}{2.1cm}
	
	\begin{pgfonlayer}{foreground}
	    \node[distlabel,anchor=west] (kittiDist) at ($(kittiZoomPos)+(.3cm,-0.05cm)$) {32\,m};
	    \draw[white,-{Latex[length=2mm,width=1.4mm]},line width=1.2pt]
	        (kittiDist.west) -- ($(kittiZoomPos)+(-0.05cm,-0.05cm)$);

	    \node[distlabel,anchor=west] (kittiDist2) at ($(kittiZoomPos)+(.05cm,0.3cm)$) {58\,m};
	    \draw[white,-{Latex[length=2mm,width=1.4mm]},line width=1.2pt]
	        (kittiDist2.north) -- ($(kittiZoomPos)+(0.55cm,0.65)$);
	        
	    \node[distlabel,anchor=west] (kittiDist3) at ($(kittiNear)+(.4cm,-0.0cm)$) {5\,m};
	    \draw[white,-{Latex[length=2mm,width=1.4mm]},line width=1.2pt]
	        (kittiDist3.west) -- ($(kittiNear)+(-0.0cm,-0.0cm)$);

	    \node[distlabelX,anchor=west] (carDist) at ($(carZoomPos1)+(0.0cm,-0.15cm)$) {21\,m};
	    \draw[white,-{Latex[length=2mm,width=1.4mm]},line width=1.2pt]
	        (carDist.west) -- ($(carZoomPos1)+(-0.42cm,-0.15cm)$);
	        
	    \node[distlabelX,anchor=west] (carDist2) at ($(carZoomPos1)+(-.8cm,0.5cm)$) {35\,m};
	    \draw[gray,-{Latex[length=2mm,width=1.4mm]},line width=1.2pt]
	        (carDist2.east) -- ($(carZoomPos1)+(0.1cm,0.5)$);
	        
	    \node[distlabelX,anchor=west] (carDist3) at ($(carZoomPos2)+(-0.4cm,0.8cm)$) {67\,m};
	    \draw[white,-{Latex[length=2mm,width=1.4mm]},line width=1.2pt]
	        (carDist3.south) -- ($(carZoomPos2)+(-0.1cm,.35cm)$);
	        
	    \node[distlabel,anchor=west] (carDist4) at ($(carZoomPos2)+(-.8cm,-0.5cm)$) {38\,m};
	    \draw[white,-{Latex[length=2mm,width=1.4mm]},line width=1.2pt]
	        (carDist4.east) -- ($(carZoomPos2)+(0.15cm,-0.5)$);
	\end{pgfonlayer}
	
	\node[label] at (r1c1.south west) {(a) KITTI};
	\node[label] at (r1c2.south west) {(b) Car};

    \coordinate (row2) at ($(r1c1.south west |- r1c2.south west)+(0,-\rowgap)$);
    \node[img, anchor=north west] (r2c1) at (row2) {\includegraphics[width=\qualimgw,trim=130px 750px 100px 350px,clip]{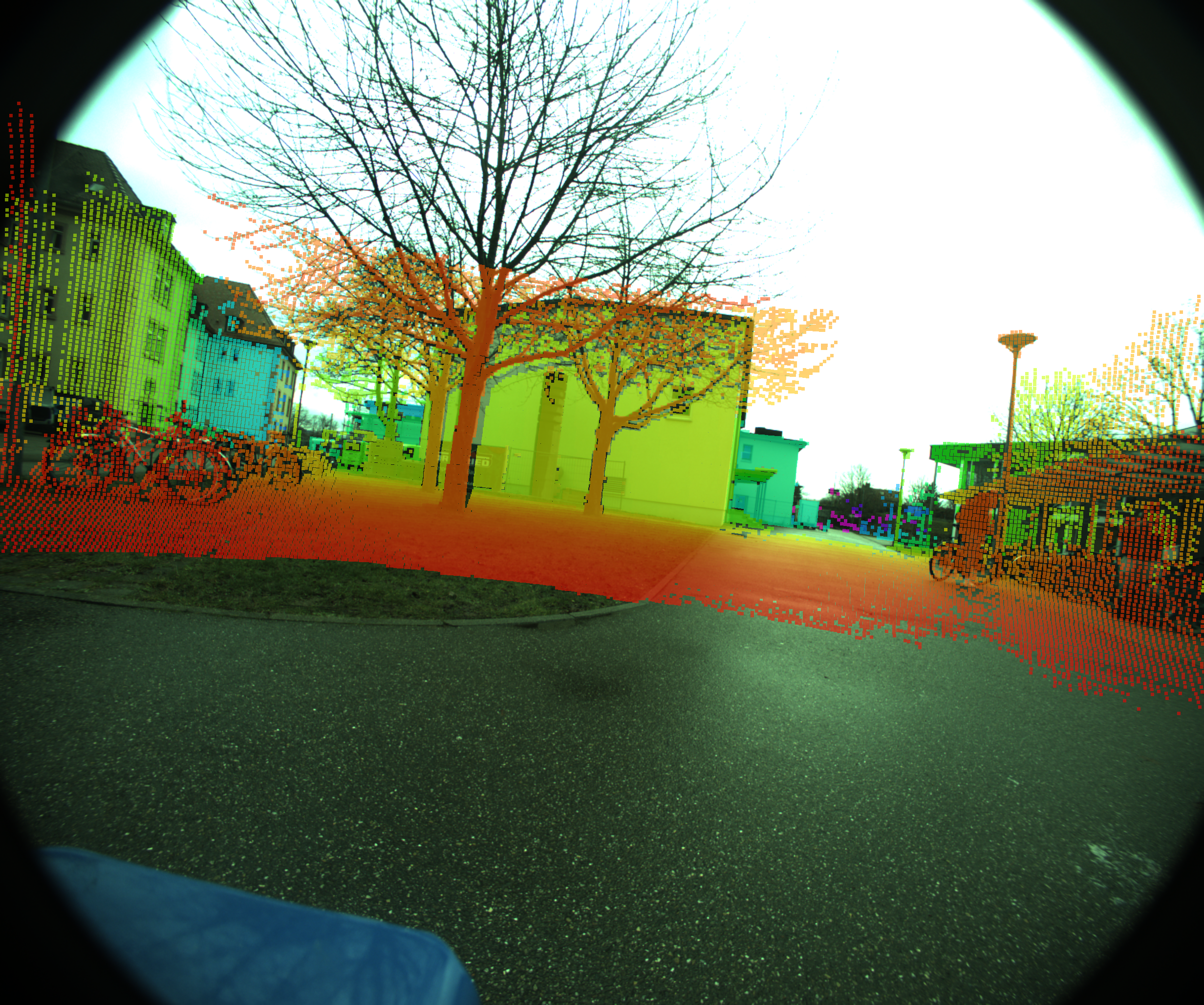}};
    \node[img, anchor=north west] (r2c2) at ($(r2c1.north east)+(\colgap,0)$) {\includegraphics[width=\qualimgw,trim=0 410 0 250px,clip]{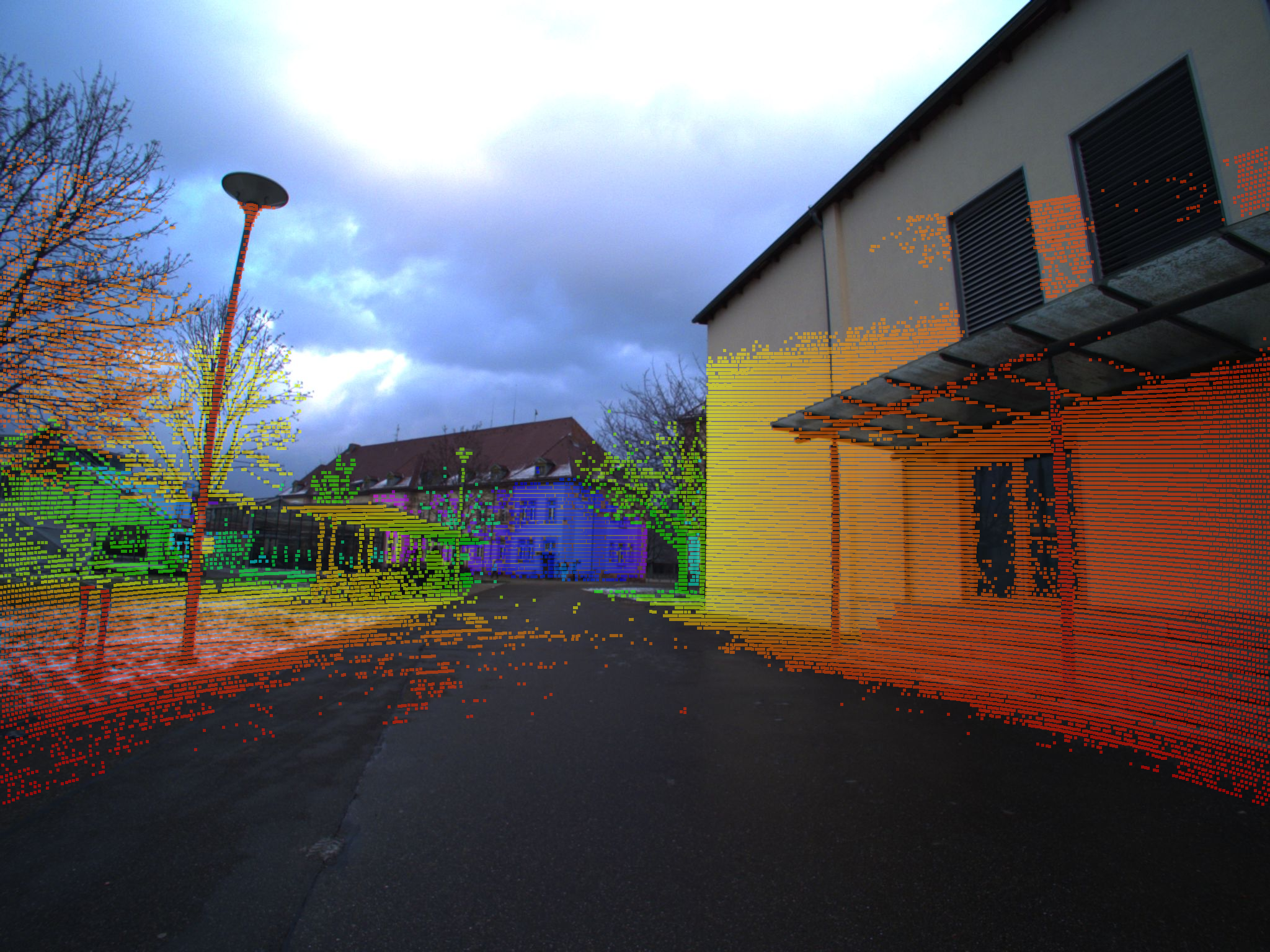}};
	
	\relcoord{r2c1}{spotNear}{0.87}{0.42}
	\draw[nearhighlight] (spotNear) ellipse[x radius=0.35cm,y radius=1.20cm];
	
	\relcoord{r2c2}{droneNear}{0.18}{0.5}
	\draw[nearhighlight] (droneNear) ellipse[x radius=0.85cm,y radius=1.85cm];
	
	\relcoord{r2c1}{spotFar1}{0.21}{0.58}
	\relcoord{r2c1}{spotZoomPos1}{0.3}{0.3}
	\zoomspy{spotFar1}{spotZoomPos1}{3.3}{2.1cm}

	\relcoord{r2c2}{droneFar1}{0.39}{0.4}
	\relcoord{r2c2}{droneZoomPos1}{0.7}{0.34}
	\zoomspy{droneFar1}{droneZoomPos1}{3.}{2.5cm}
	
	\begin{pgfonlayer}{foreground}
	    \node[distlabelX,anchor=west] (spotDist) at ($(spotZoomPos1)+(-.5cm,0.5cm)$) {20\,m};
	    \draw[gray,-{Latex[length=2mm,width=1.4mm]},line width=1.2pt]
	        (spotDist.east) -- ($(spotZoomPos1)+(0.43cm,0.5cm)$);
	        
	    \node[distlabelX,anchor=west] (spotDist2) at ($(spotZoomPos1)+(.05cm,0.1cm)$) {48\,m};
	    \draw[white,-{Latex[length=2mm,width=1.4mm]},line width=1.2pt]
	        (spotDist2.west) -- ($(spotZoomPos1)+(-0.36cm,0.1)$);
	        
	    \node[distlabel,anchor=west] (spotDist3) at ($(spotNear)+(.38cm,-0.0cm)$) {10\,m};
	    \draw[white,-{Latex[length=2mm,width=1.4mm]},line width=1.2pt]
	        (spotDist3.west) -- ($(spotNear)+(-0.05cm,-0.0cm)$);

	    \node[distlabel,anchor=west] (droneDist) at ($(droneZoomPos1)+(-.3cm,0.05cm)$) {24\,m};
	    \draw[white,-{Latex[length=2mm,width=1.4mm]},line width=1.2pt]
	        (droneDist.west) -- ($(droneZoomPos1)+(-0.65cm,0.05cm)$);
	        
	    \node[distlabel,anchor=west] (droneDist2) at ($(droneZoomPos1)+(.25cm,0.45cm)$) {62\,m};
	    \draw[white,-{Latex[length=2mm,width=1.4mm]},line width=1.2pt]
	        (droneDist2.south) -- ($(droneZoomPos1)+(0.55cm,-0.02)$);
	        
	    \node[distlabel,anchor=west] (droneDist3) at ($(droneNear)+(.3cm,-0.0cm)$) {6\,m};
	    \draw[white,-{Latex[length=2mm,width=1.4mm]},line width=1.2pt]
	        (droneDist3.west) -- ($(droneNear)+(-0.13cm,-0.0cm)$);
	\end{pgfonlayer}
	
	\node[label] at (r2c1.south west) {(c) Spot};
	\node[label] at (r2c2.south west) {(d) UAV};

    \end{tikzpicture}
	\vspace{-.7em}
    \caption{Qualitative calibration results on KITTI and the three robotic platforms. Blue colored near-field highlights emphasize structures near the image borders, where the camera-LiDAR projection is particularly sensitive to intrinsic and distortion parameters, demonstrating that our method recovers them accurately. Circular zoom-ins show distant structures at the annotated ranges, demonstrating precise joint calibration over long distances.}
    \label{fig:qualitative_results}
\end{figure*}

We further evaluate our method on three real robotic platforms with different cameras, LiDAR sensors, and mounting configurations (cf. Fig.~\ref{fig:teaser}, top).
For our method, COLMAP provides the initial camera parameters, while ORB-SLAM3 and FAST-LIO2 provide visual and LiDAR trajectories for automatic extrinsic initialization, yielding a fully practical setting. %
To evaluate robustness to imperfect camera initialization, we use the perturbed COL$^+$ setting. 
Tab.~\ref{tab:robot_results} compares our approach with the target-less methods of Koide~et~al.~\cite{koide_general_2023}, MDPCalib~\cite{petek_automatic_2024}, and Shen~et~al.~\cite{shen_direct_2025}. The first two are extrinsic-only methods and assume known, fixed camera intrinsics.
For the joint method~\cite{shen_direct_2025}, we use perturbed COLMAP intrinsics as in our setting, but retain the original CAD-style extrinsic initialization, as the method did not reliably converge from the larger offsets of our automatic initializer.
Despite this more favorable initialization, the baseline mainly improves rotation, while translation and reprojection errors remain large. 
We observe that our method achieves the lowest joint-calibration errors on all three platforms while also recovering the camera parameters. On Car and Spot, it even outperforms the extrinsic-only methods using fixed ground-truth intrinsics, while on the UAV only Koide~et~al.~\cite{koide_general_2023} retain a lower extrinsics error.

\begin{table}[t]
\centering
\caption{Target-less calibration results on real robotic platforms.} \vspace{-1em}
\label{tab:robot_results}
\setlength{\tabcolsep}{1.6pt}
\renewcommand{\arraystretch}{1.05}
\begin{tabular}{lllccc}
\toprule
Robot
&
Method
&
Setting
&
Trans. [cm]
&
Rot. [$^\circ$]
&
Reproj. [px]
\\
\midrule
\multirow{4}{*}{Car}
&
Koide~\cite{koide_general_2023}
&
fixed GT
&
9.76 & 0.27 & --
\\
&
MDP~\cite{petek_automatic_2024}
&
fixed GT
&
4.50 & 0.27 & --
\\
\cmidrule(lr){2-6}
&
Shen~\cite{shen_direct_2025}
&
opt. COL$^+$
&
$20.0\pm13.1$ & $1.24\pm0.65$ & $46.6\pm72.0$ 
\\
&
Ours
&
opt. COL$^+$
&
$\mathbf{3.02}\pm\mathbf{0.11}$ & $\mathbf{0.09}\pm\mathbf{0.04}$ & $\mathbf{3.63}\pm\mathbf{0.48}$
\\
\midrule
\multirow{4}{*}{Spot}
&
Koide~\cite{koide_general_2023}
&
fixed GT
&
16.21 & 1.34 & --
\\
&
MDP~\cite{petek_automatic_2024}
&
fixed GT
&
9.54 & 0.38 & --
\\
\cmidrule(lr){2-6}
&
Shen~\cite{shen_direct_2025}
&
opt. COL$^+$
&
$30.0\pm23.9$ & $1.69\pm1.03$ & $41.3\pm21.4$ 
\\
&
Ours
&
opt. COL$^+$
&
$\mathbf{2.18}\pm\mathbf{0.40}$ & $\mathbf{0.28}\pm\mathbf{0.02}$ & $\mathbf{9.21}\pm\mathbf{0.17}$ %
\\
\midrule
\multirow{4}{*}{UAV}
&
Koide~\cite{koide_general_2023}
&
fixed GT
&
1.65 & 0.36 & --
\\
&
MDP~\cite{petek_automatic_2024}
&
fixed GT
&
5.12 & 0.51 & --
\\
\cmidrule(lr){2-6}
&
Shen~\cite{shen_direct_2025}
&
opt. COL$^+$
&
$10.3\pm8.93$ & $0.90\pm0.40$ & $40.7\pm87.8$ 
\\
&
Ours
&
opt. COL$^+$
&
$\mathbf{1.91}\pm\mathbf{0.22}$ & $\mathbf{0.48}\pm\mathbf{0.01}$ & $\mathbf{2.00}\pm\mathbf{0.09}$ 
\\
\bottomrule
\end{tabular}

\vspace{0.3em}
\footnotesize
\raggedright
Koide~et~al. \& MDPCalib results from~\cite{petek_automatic_2024} using fixed reference intrinsics.
Joint results report mean$\pm$std over 100 trials. Bold denotes best joint result.
\end{table}

\subsection{Qualitative Results}

\begin{table}
\centering
\caption{Runtime Evaluation. Mean $\pm$ std over 100 runs.}
\vspace{-1em}
\label{tab:runtime}
\setlength{\tabcolsep}{2pt}
\renewcommand{\arraystretch}{1.05}
\begin{tabular}{lccc}
\toprule
Stage & Inference [s] & Optimization [s] & Total [s] \\
\midrule
$j \leq N_{\mathrm{intr}}$, per it.
& $16.94 \pm 0.09$
& $15.32 \pm 5.05$
& $32.25 \pm 5.05$ \\
$j > N_{\mathrm{intr}}$, per it.
& $16.94 \pm 0.05$
& $\phantom{0}5.10 \pm 1.32$
& $22.04 \pm 1.32$ \\
\midrule
Complete calibration
& $186.35 \pm 0.66$
& $107.19 \pm 10.80$
& $302.13 \pm 16.58$ \\
\bottomrule
\end{tabular}

\vspace{0.3em}
\footnotesize
\raggedright
per it.: per iteration, comprising 100 image-scan pairs.
\end{table}

Fig.~\ref{fig:qualitative_results} presents qualitative reprojection results for the final joint calibration on KITTI and the three real robotic platforms. The projected LiDAR structures align closely with image edges across a wide range of depths. In particular, structures near the image borders remain well aligned despite increased sensitivity to the intrinsic and distortion parameters, while the zoomed distant regions show accurate camera-LiDAR alignment over long ranges.

\subsection{Runtime}

Tab.~\ref{tab:runtime} summarizes the runtime of the iterative calibration procedure, with each iteration processing 100 image-scan pairs. Since optimizing the camera parameters increases the nonlinear optimization cost, we report iterations with active intrinsic refinement, $j \leq N_{\mathrm{intr}}$, separately from the subsequent pose-only iterations, $j > N_{\mathrm{intr}}$. Per-iteration totals are the sum of correspondence inference and refinement optimization. For the complete calibration, the total corresponds to the measured wall time and therefore additionally includes initialization, data handling, and other overhead.
All runtimes were measured on a workstation with an AMD Ryzen 9 PRO 7945 CPU (24 threads), 64\,GB RAM, and an NVIDIA GeForce RTX 4060 Ti (16\,GB VRAM, approx. 3.5\,GB used by the correspondence network at the KITTI resolution of $1392\times512$). 
With the default $N_{\mathrm{total}}=11$ and $N_{\mathrm{intr}}=5$, the complete calibration requires approx. $300\,\mathrm{s}$, thereof about $62\%$ for correspondence inference, $35\%$ for nonlinear optimization, and $3\%$ for other overhead.
In comparison, MDPCalib~\cite{petek_automatic_2024} reports approx. $3.5\times10^{3}\,\mathrm{s}$ for its evaluated setting.

\section{Conclusion}

We presented a target-less framework for joint intrinsic and extrinsic camera-LiDAR calibration based on learned pixel-point correspondences and geometric optimization. By making correspondence learning aware of camera intrinsics and distortion, the method operates directly on raw images and generalizes to unseen camera-LiDAR configurations. Alternating correspondence prediction with joint nonlinear refinement recovers accurate camera parameters while achieving lower extrinsic errors than prior extrinsic-only methods that assume known intrinsics. Experiments on KITTI and multiple robotic platforms demonstrate practical calibration from COLMAP-based initialization without privileged calibration information. Future work will investigate automatic miscalibration detection and online recalibration.

\bibliographystyle{IEEEtran}
\bibliography{references.bib}

\begin{thebibliography}{10}
\providecommand{\url}[1]{#1}
\csname url@rmstyle\endcsname
\providecommand{\newblock}{\relax}
\providecommand{\bibinfo}[2]{#2}
\providecommand\BIBentrySTDinterwordspacing{\spaceskip=0pt\relax}
\providecommand\BIBentryALTinterwordstretchfactor{4}
\providecommand\BIBentryALTinterwordspacing{\spaceskip=\fontdimen2\font plus
\BIBentryALTinterwordstretchfactor\fontdimen3\font minus
  \fontdimen4\font\relax}
\providecommand\BIBforeignlanguage[2]{{%
\expandafter\ifx\csname l@#1\endcsname\relax
\typeout{** WARNING: IEEEtran.bst: No hyphenation pattern has been}%
\typeout{** loaded for the language `#1'. Using the pattern for}%
\typeout{** the default language instead.}%
\else
\language=\csname l@#1\endcsname
\fi
#2}}

\bibitem{valada2016towards}
A.~Valada, G.~Oliveira, T.~Brox, and W.~Burgard, ``Towards robust semantic
  segmentation using deep fusion,'' in \emph{Robot.: Sci. Syst. (RSS) Workshop
  Are the Sceptics Right? Limits Potentials Deep Learn. Robot.}, vol. 114,
  2016.

\bibitem{bultmann2021real}
S.~Bultmann, J.~Quenzel, and S.~Behnke, ``Real-time multi-modal semantic fusion
  on unmanned aerial vehicles,'' in \emph{Eur. Conf. Mobile Robots (ECMR)},
  2021.

\bibitem{mohan2026up}
R.~Mohan, F.~Drews, Y.~Miron, D.~Cattaneo, and A.~Valada, ``{UP-Fuse}:
  Uncertainty-guided {LiDAR}-camera fusion for {3D} panoptic segmentation,'' in
  \emph{Robot.: Sci. Syst. (RSS)}, 2026.

\bibitem{hindel2026cotrate}
J.~Hindel, S.~Bultmann, H.~Masnavi, D.~Cattaneo, and A.~Valada,
  ``Self-supervised online robot-agnostic traversability estimation for
  open-world environments,'' \emph{arXiv:2605.28442}, 2026.

\bibitem{rehder_kalibr_2016}
J.~Rehder, J.~Nikolic, T.~Schneider, T.~Hinzmann, and R.~Siegwart, ``Extending
  {Kalibr}: {Calibrating} the extrinsics of multiple {IMUs} and of individual
  axes,'' in \emph{IEEE Int. Conf. Robot. Autom. (ICRA)}, 2016, pp. 4304--4311.

\bibitem{kuemmerle_unified_2020}
J.~Kümmerle and T.~Kühner, ``Unified intrinsic and extrinsic camera and lidar
  calibration under uncertainties,'' in \emph{IEEE Int. Conf. Robot. Autom.
  (ICRA)}, 2020, pp. 6028--6034.

\bibitem{yan_joint_2023}
G.~Yan, F.~He, C.~Shi, P.~Wei, X.~Cai, and Y.~Li, ``Joint camera intrinsic and
  {LiDAR}-camera extrinsic calibration,'' in \emph{IEEE Int. Conf. Robot.
  Autom. (ICRA)}, 2023, pp. 11\,446--11\,452.

\bibitem{wiesmann_joint_2024}
L.~Wiesmann, T.~Läbe, L.~Nunes, J.~Behley, and C.~Stachniss, ``Joint intrinsic
  and extrinsic calibration of perception systems utilizing a calibration
  environment,'' \emph{IEEE Robot. Autom. Lett.}, vol.~9, no.~10, pp.
  9103--9110, 2024.

\bibitem{paetzold_camcalib_2022}
B.~P{\"a}tzold, S.~Bultmann, and S.~Behnke, ``Online marker-free extrinsic
  camera calibration using person keypoint detections,'' in \emph{DAGM German
  Conf. Pattern Recognit. (GCPR)}, 2022, pp. 300--316.

\bibitem{li_joint_2023}
L.~Li, H.~Li, X.~Liu, D.~He, Z.~Miao, F.~Kong, R.~Li, Z.~Liu, and F.~Zhang,
  ``Joint intrinsic and extrinsic {LiDAR}-camera calibration in targetless
  environments using plane-constrained bundle adjustment,''
  \emph{arXiv:2308.12629}, 2023.

\bibitem{petek_automatic_2024}
K.~Petek, N.~Vödisch, J.~Meyer, D.~Cattaneo, A.~Valada, and W.~Burgard,
  ``Automatic target-less camera-{LiDAR} calibration from motion and deep point
  correspondences,'' \emph{IEEE Robot. Autom. Lett.}, vol.~9, no.~11, pp.
  9978--9985, 2024.

\bibitem{cattaneo2020cmrnetpp}
D.~Cattaneo, D.~G. Sorrenti, and A.~Valada, ``Cmrnet++: Map and camera agnostic
  monocular visual localization in lidar maps,'' in \emph{IEEE Int. Conf.
  Robot. Autom. (ICRA) Workshop Emerging Learn. Algorithmic Methods Data Assoc.
  Robot.}, 2020.

\bibitem{cattaneo_cmrnext_2025}
D.~Cattaneo and A.~Valada, ``{CMRNext}: Camera to {LiDAR} matching in the wild
  for localization and extrinsic calibration,'' \emph{IEEE Trans. Robot.},
  vol.~41, pp. 1995--2013, 2025.

\bibitem{nayak2024ralf}
A.~Nayak, D.~Cattaneo, and A.~Valada, ``{RaLF}: Flow-based global and metric
  radar localization in {LiDAR} maps,'' in \emph{IEEE Int. Conf. Robot. Autom.
  (ICRA)}, 2024, pp. 5097--5103.

\bibitem{yu_i2dloc_2024}
H.~Yu, K.~Chen, W.~Yang, S.~Scherer, and G.-S. Xia, ``{I2D-Loc++}: Camera pose
  tracking in {LiDAR} maps with multi-view motion flows,'' \emph{IEEE Robot.
  Autom. Lett.}, vol.~9, no.~9, pp. 8162--8169, 2024.

\bibitem{huang_what_2025}
S.~Huang, C.~Lin, and Y.~Zhao, ``What really matters for learning-based
  lidar-camera calibration,'' in \emph{ACM Int. Conf. Multimedia Asia
  (MMAsia)}, 2025.

\bibitem{yuan_pixel_2021}
C.~Yuan, X.~Liu, X.~Hong, and F.~Zhang, ``Pixel-level extrinsic self
  calibration of high resolution {LiDAR} and camera in targetless
  environments,'' \emph{IEEE Robot. Autom. Lett.}, vol.~6, no.~4, pp.
  7517--7524, 2021.

\bibitem{yin_automatic_2023}
J.~Yin, F.~Yan, Y.~Liu, and Y.~Zhuang, ``Automatic and targetless
  {LiDAR}--camera extrinsic calibration using edge alignment,'' \emph{IEEE
  Sens. J.}, vol.~23, no.~17, pp. 19\,871--19\,880, 2023.

\bibitem{koide_general_2023}
K.~Koide, S.~Oishi, M.~Yokozuka, and A.~Banno, ``General, single-shot,
  target-less, and automatic {LiDAR}-camera extrinsic calibration toolbox,'' in
  \emph{IEEE Int. Conf. Robot. Autom. (ICRA)}, 2023, pp. 11\,301--11\,307.

\bibitem{schneider_regnet_2017}
N.~Schneider, F.~Piewak, C.~Stiller, and U.~Franke, ``{RegNet}: Multimodal
  sensor registration using deep neural networks,'' in \emph{IEEE Intell. Veh.
  Symp. (IV)}, 2017, pp. 1803--1810.

\bibitem{iyer_calibnet_2018}
G.~Iyer, R.~K. Ram, J.~K. Murthy, and K.~M. Krishna, ``{CalibNet}:
  Geometrically supervised extrinsic calibration using 3d spatial transformer
  networks,'' in \emph{IEEE/RSJ Int. Conf. Intell. Robots Syst. (IROS)}, 2018,
  pp. 1110--1117.

\bibitem{lv_lccnet_2021}
X.~Lv, B.~Wang, Z.~Dou, D.~Ye, and S.~Wang, ``{LCCNet}: {LiDAR} and camera
  self-calibration using cost volume network,'' in \emph{IEEE/CVF Conf. Comput.
  Vis. Pattern Recognit. Workshops (CVPRW)}, 2021, pp. 2888--2895.

\bibitem{herau_moisst_2023}
Q.~Herau, N.~Piasco, M.~Bennehar, L.~Roldão, D.~Tsishkou, C.~Migniot,
  P.~Vasseur, and C.~Demonceaux, ``{MOISST}: Multimodal optimization of
  implicit scene for spatiotemporal calibration,'' in \emph{IEEE/RSJ Int. Conf.
  Intell. Robots Syst. (IROS)}, 2023, pp. 1810--1817.

\bibitem{herau_soac_2024}
Q.~Herau, N.~Piasco, M.~Bennehar, L.~Roldao, D.~Tsishkou, C.~Migniot,
  P.~Vasseur, and C.~Demonceaux, ``{SOAC}: Spatio-temporal overlap-aware
  multi-sensor calibration using neural radiance fields,'' in \emph{IEEE/CVF
  Conf. Comput. Vis. Pattern Recognit. (CVPR)}, 2024, pp. 15\,131--15\,140.

\bibitem{herau_3dgscalib_2024}
Q.~Herau, M.~Bennehar, A.~Moreau, N.~Piasco, L.~Roldão, D.~Tsishkou,
  C.~Migniot, P.~Vasseur, and C.~Demonceaux, ``{3DGS-Calib}: {3D} gaussian
  splatting for multimodal spatiotemporal calibration,'' in \emph{IEEE/RSJ Int.
  Conf. Intell. Robots Syst. (IROS)}, 2024, pp. 8315--8321.

\bibitem{zhou_robust_2025}
S.~Zhou, S.~Xie, R.~Ishikawa, and T.~Oishi, ``Robust {LiDAR}-camera calibration
  with {2D} gaussian splatting,'' \emph{IEEE Robot. Autom. Lett.}, vol.~10,
  no.~5, pp. 4674--4681, 2025.

\bibitem{jung_targetless_2025}
H.~Jung, N.~Kim, J.~Kim, and J.~Park, ``Targetless lidar-camera calibration
  with neural gaussian splatting,'' \emph{IEEE Robot. Autom. Lett.}, vol.~11,
  no.~4, pp. 4777--4784, 2026.

\bibitem{kitti_geiger_2012}
A.~Geiger, P.~Lenz, and R.~Urtasun, ``Are we ready for autonomous driving? the
  {KITTI} vision benchmark suite,'' in \emph{IEEE Conf. Comput. Vis. Pattern
  Recognit. (CVPR)}, 2012, pp. 3354--3361.

\bibitem{shen_direct_2025}
Y.~Shen, S.~Hong, S.~Shen, and T.~Qin, ``Direct, targetless and automatic joint
  calibration of {LiDAR}-camera intrinsic and extrinsic,'' in \emph{IEEE/RSJ
  Int. Conf. Intell. Robots Syst. (IROS)}, 2025, pp. 8720--8726.

\bibitem{leotta_maximum_2022}
M.~J. Leotta, D.~Russell, and A.~Matrai, ``On the maximum radius of polynomial
  lens distortion,'' in \emph{IEEE/CVF Winter Conf. Appl. Comput. Vis. (WACV)},
  2022, pp. 2374--2382.

\bibitem{Agarwal_Ceres_Solver_2022}
\BIBentryALTinterwordspacing
S.~Agarwal, K.~Mierle, and T.~C.~S. Team, ``{Ceres Solver},'' 2023. [Online].
  Available: \url{https://github.com/ceres-solver/ceres-solver}
\BIBentrySTDinterwordspacing

\bibitem{schoenberger2016sfm}
J.~L. Sch\"{o}nberger and J.-M. Frahm, ``Structure-from-motion revisited,'' in
  \emph{IEEE/CVF Conf. Comput. Vis. Pattern Recognit. (CVPR)}, 2016, pp.
  4104--4113.

\bibitem{ORBSLAM3_TRO}
C.~Campos, R.~Elvira, J.~J.~G. Rodríguez, J.~M. M.~Montiel, and J.~D.~Tardós,
  ``{ORB-SLAM3}: An accurate open-source library for visual, visual-inertial
  and multi-map {SLAM},'' \emph{IEEE Trans. Robot.}, vol.~37, no.~6, pp.
  1874--1890, 2021.

\bibitem{xu_fastlio2_2022}
W.~Xu, Y.~Cai, D.~He, J.~Lin, and F.~Zhang, ``{FAST-LIO2}: Fast direct
  {LiDAR}-inertial odometry,'' \emph{IEEE Trans. Robot.}, vol.~38, no.~4, pp.
  2053--2073, 2022.

\bibitem{argoverse_chang_2019}
M.-F. Chang, J.~Lambert, P.~Sangkloy, J.~Singh, S.~Bak, A.~Hartnett, D.~Wang,
  P.~Carr, S.~Lucey, D.~Ramanan, and J.~Hays, ``Argoverse: {3D} tracking and
  forecasting with rich maps,'' in \emph{IEEE/CVF Conf. Comput. Vis. Pattern
  Recognit. (CVPR)}, 2019, pp. 8740--8749.

\bibitem{pandaset_xiao_2021}
P.~Xiao, Z.~Shao, S.~Hao, Z.~Zhang, X.~Chai, J.~Jiao, Z.~Li, J.~Wu, K.~Sun,
  K.~Jiang, Y.~Wang, and D.~Yang, ``{PandaSet}: Advanced sensor suite dataset
  for autonomous driving,'' in \emph{IEEE Intell. Transp. Syst. Conf. (ITSC)},
  2021, pp. 3095--3101.

\end{thebibliography}

\end{document}